\documentclass[journal]{IEEEtran}

\usepackage{graphics} %
\usepackage{graphicx}
\usepackage{algpseudocode,algorithm}

\usepackage{amsmath,amsfonts,amssymb}
\usepackage{color}
\usepackage{wrapfig}
\usepackage{subfigure}
\usepackage{multirow}
\usepackage{array}
\usepackage[table]{xcolor}
\usepackage{soul}
\usepackage{mathtools}
\usepackage{fixmath}
\usepackage{nccmath}
\usepackage{cuted}
\usepackage{float}
\usepackage{tikz}
\newcolumntype{P}[1]{>{\centering\arraybackslash}p{#1}}
\newcolumntype{M}[1]{>{\centering\arraybackslash}m{#1}}

\DeclareMathOperator*{\argmin}{arg\,min}
\newcommand{\etal}{\textit{et al}.}
\newcommand{\ie}{\textit{i}.\textit{e}.}
\newcommand{\eg}{\textit{e}.\textit{g}.}

\def\checkmark{\tikz\fill[scale=0.4](0,.35) -- (.25,0) -- (1,.7) -- (.25,.15) -- cycle;} 
\newcolumntype{L}[1]{>{\raggedright\let\newline\\\arraybackslash\hspace{0pt}}m{#1}}
\newcolumntype{C}[1]{>{\centering\let\newline\\\arraybackslash\hspace{0pt}}m{#1}}
\newcolumntype{R}[1]{>{\raggedleft\let\newline\\\arraybackslash\hspace{0pt}}m{#1}}

\begin{document}
\title{Canny-VO: Visual Odometry with RGB-D Cameras based on Geometric 3D-2D Edge Alignment}

\author{Yi~Zhou,~\IEEEmembership{Student Member,~IEEE,}
		Hongdong Li,~\IEEEmembership{Member,~IEEE} and 		Laurent~Kneip,~\IEEEmembership{Member,~IEEE}
\thanks{Yi Zhou and Hongdong Li are with the Research School of Engineering, the Australian National University. E-mail: {yi.zhou, hongdong.li}@anu.edu.au. Laurent Kneip is with the School of Information Science and Technology, ShanghaiTech. Email: lkneip@shanghaitech.edu.cn.}}

\markboth{IEEE TRANSACTIONS ON ROBOTICS,~Vol.~, No.~, Month~Year}%
{Shell \MakeLowercase{\textit{et al.}}: Bare Demo of IEEEtran.cls for IEEE Journals}

\maketitle
\begin{abstract}
The present paper reviews the classical problem of free-form curve registration and applies it to an efficient RGB-D visual odometry system called Canny-VO, as it efficiently tracks all Canny edge features extracted from the images. Two replacements for the distance transformation commonly used in edge registration are proposed: Approximate Nearest Neighbour Fields and Oriented Nearest Neighbour Fields. 3D-2D edge alignment benefits from these alternative formulations in terms of both efficiency and accuracy. It removes the need for the more computationally demanding paradigms of data-to-model registration, bilinear interpolation, and sub-gradient computation. To ensure robustness of the system in the presence of outliers and sensor noise, the registration is formulated as a maximum a posteriori problem, and the resulting weighted least squares objective is solved by the iteratively re-weighted least squares method. A variety of robust weight functions are investigated and the optimal choice is made based on the statistics of the residual errors. Efficiency is furthermore boosted by an adaptively sampled definition of the nearest neighbour fields. Extensive evaluations on public SLAM benchmark sequences demonstrate state-of-the-art performance and an advantage over classical Euclidean distance fields.
\end{abstract}


\IEEEpeerreviewmaketitle

\section{Introduction}
\IEEEPARstart{I}{mage}-based estimation of camera motion---known as visual odometry (VO)---plays an important role in many applications such as control and navigation of unmanned mobile robots, especially when no external reference signal is available. Over the past decade, we have witnessed a number of successful works, such as salient feature based sparse methods~\cite{klein2007parallel, mur2015orb}, direct methods~\cite{tykkala2011direct,steinbrucker2011real,audras2011real,kerl2013robust} that employ all intensity information in the image, semi-dense pipelines~\cite{engel2013semi,engel2014lsd} and other systems like ~\cite{newcombe2011kinectfusion,whelan2012kintinuous,pomerleau2011tracking,pomerleau2013comparing} which track the camera using an ICP algorithm over the depth information. The present work focusses on edge-based registration, which finds a good compromise between the amount of data used for registration and computational complexity.

Considering that edge detectors have been discovered before invariant keypoint extractors, it comes as no surprise that pioneering works in computer vision such as Larry Robert's idea of a \textit{block's world} \cite{roberts65} envisage the mapping and registration of entire 3D curves rather than ``just" sparse 3D points. While sparse point-based methods have proven to be very effective at subtracting the correspondence problem from the inverse problem of structure from motion, curve-based estimation remains interesting due to the following, geometrically motivated advantages:
\begin{itemize}
\item Edges in images make up for a significantly larger amount of data points to be registered to a model, hence leading to superior signal-to-noise ratio and improved overall accuracy.
\item Edges represent a more natural choice in man-made environments, where objects are often made up of homogeneously coloured (i.e. texture-less) piece-wise planar surfaces.
\item Lines and curves lead to more meaningful 3D representations of the environment than points. Curve-based 3D models may for instance ease the inference of object shapes, sizes and boundaries.
\end{itemize}

It is the correspondence problem and the resulting computational complexity which however prevented practical, edge or curve-based tracking and mapping pipelines from appearing in the literature until only very recently. Knowing which point from a 3D curve reprojects to which point from a 2D curve measured in the image plane is only easy once the registration problem is solved. Therefore, the correspondence problem has to be solved as part of the 3D-2D registration. Research around the iterative closest point paradigm \cite{chen1992object}, distance transformations \cite{kimmel96}, and more recent advances such as continuous spline-based parametrisations \cite{xiao05,nurutdinova2015towards} nowadays alleviate the iterative computation of putative correspondences, thus rendering online free-form curve-based registration possible.

The contributions of this paper read as follows:

\begin{itemize}
\item A detailed review of 3D-2D free-form edge alignment, summarizing the difficulties of the problem and the solutions given by existing real-time edge alignment methods in robotics.
\item Two alternatives to distance transformations --- \textsl{Approximate Nearest Neighbour Fields} and \textsl{Oriented Nearest Neighbour Fields} --- with properties that improve the registration in terms of efficiency and accuracy.
\item A real-time RGB-D visual odometry system based on nearest neighbour fields, which achieves robust tracking by formulating the 3D-2D ICP based motion estimation as a maximum a posteriori problem.
\item An extensive evaluation on publicly available RGB-D datasets and a performance comparison that demonstrates the improvements over previous state-of-the-art edge alignment methods.
\end{itemize}
The paper is organized as follows. 
More related work is discussed in Section~\ref{Sec: Related work}. 
Section~\ref{Sec: Core concept} provides a review of geometric 3D-2D edge alignment, the problems resulting from employing Euclidean distance fields, and the corresponding solutions of existing methods. 
Sections~\ref{Sec:Approximate Nearest Neighbour Fields} and~\ref{sec:oriented nearest neighbour fields} detail our novel distance transformation alternatives --- \textit{Approximate Nearest Neighbour Fields} and \textit{Oriented Nearest Neighbour Fields}.
Section~\ref{sec: robust vo system} outlines our complete Canny-VO system with an emphasis on robust weighting for accurate motion estimation in the presence of noise and outliers.
Section~\ref{Sec: Evaluation} concludes with our extensive experimental evaluation.

\section{Related Work}
\label{Sec: Related work}

Curve-based structure from motion has a long-standing tradition in geometric computer vision. Early work by Porrill and Pollard \cite{porrill91} has discovered how curve and surface tangents can be included into fundamental epipolar geometry for stereo calibration, an idea later on followed up by Feldmar \etal ~\cite{feldmar95} and Kaminski and Shashua \cite{kaminski04}. However, the investigated algebraic constraints for solving multiple view geometry problems are known to be very easily affected by noise. In order to improve the quality of curve-based structure from motion, further works by Faugeras and Mourrain \cite{faugeras95} and Kahl and Heyden \cite{kahl98} therefore looked at special types of curves such as straight lines and cones, respectively.

In contrast to those early contributions in algebraic geometry, a different line of research is formed by works that investigate curve-based structure from motion from the point of view of 3D model parametrisation and optimisation. Kahl and August \cite{kahl03} are among the first to show complete, free-form 3D curve reconstruction from registered 2D images. Later works then focus on improving the parametrisation of the 3D curves, presenting sub-division curves \cite{kaess04}, non-rational B-splines \cite{xiao05}, and implicit representations via 3D probability distributions \cite{teney12}. These works, however, mostly focus on the reconstruction problem, and do not use the curve measurements in order to refine the camera poses.

Complete structure-from-motion optimisation including general curve models and camera poses has first been shown by Berthilsson \etal ~\cite{berthilsson01}. The approach however suffers from a bias that occurs when the model is only partially observed. Nurutdinova and Fitzgibbon \cite{nurutdinova2015towards} illustrate this problem in detail, and present an inverse data-to-model registration concept that transparently handles missing data. Fabbri and Kimia \cite{fabbri10} solve the problem by modelling curves as a set of shorter line segments, and Cashman and Fitzgibbon \cite{cashman13} model the occlusions explicitly. The successful inclusion of shorter line segments (i.e. edglets) has furthermore been demonstrated in real-time visual SLAM \cite{eade2009edge}. Further related work from the visual SLAM community is given by Engel \etal ~\cite{engel2013semi,engel2014lsd}, who estimate semi-dense depth maps in high-gradient regions of the image, and then register subsequent images based on a photometric error criterion. As common with all direct photometric methods, however, the approach is difficult to combine with a global optimization of structure, and easily affected by illumination changes.

The core problem of projective 3D-to-2D free-form curve registration goes back to the difficulty of establishing correspondences in the data. The perhaps most traditional solution to this problem is given by the ICP algorithm \cite{chen1992object,besl92,pomerleau2013comparing}. Yang \etal ~\cite{yang2016goicp} even developed a globally optimal variant of the ICP algorithm, which is however too slow for most practically relevant use-cases. Pomerleau \etal ~\cite{pomerleau2011tracking} and Tykk{\"a}l{\"a} \etal ~\cite{tykkala2011direct} present real-time camera pose registration algorithms based on the ICP algorithm, where the latter work minimises a hybrid geometry and appearance based cost function. Both works however cast the alignment problem as a 3D-3D registration problem. More recently, Kneip \etal ~\cite{BMVC2015_100} show how to extend the idea to 3D-2D registration of edge-based depth maps in a reference frame.

The caveat of the ICP algorithm is given by the repetitive requirement to come up with putative correspondences that still can help to improve the registration. Zhang \cite{zhang94} investigated how this expensive search can be speeded up by pre-structuring the data in a K-D tree. The biggest leap with respect to classical ICP was however achieved through the introduction of distance fields \cite{kimmel96}. Newcombe \etal ~\cite{newcombe2011kinectfusion} and Bylow \etal ~\cite{bylow13} for instance rely on distance fields to perform accurate real-time tracking of a depth sensor. Steinbr\"ucker \etal ~\cite{steinbruecker13} furthermore push the efficiency by adaptive sampling of the distance field \cite{frisken00}. More recently, distance-field based registration has also been introduced in the context of 3D-to-2D registration. Kneip \etal ~\cite{BMVC2015_100} and Kuse and Shen \cite{kuse2016robust} show the successful use of 2D distance fields for projective registration of 3D curves. Our work follows up on this line of research, and proposes a yet more efficient alternative to distance fields for 3D-2D, ICP-based curve registration. Our oriented nearest neighbour fields notably do not suffer from the previously identified registration bias in the case of partially observed models.

\section{Review of Geometric 3D-2D Edge Registration}
\label{Sec: Core concept}

This section reviews the basic idea behind geometric 3D-2D curve alignment. After a clear problem definition, we discuss the limitations of existing Euclidean distance-field based methods addressed through our work.
\subsection{Problem statement}
\label{Subsec: Problem statement}

\begin{figure}[b]
  \centering
  \subfigure[Image gradient's norm map.]{
  \includegraphics[width=0.45\columnwidth]{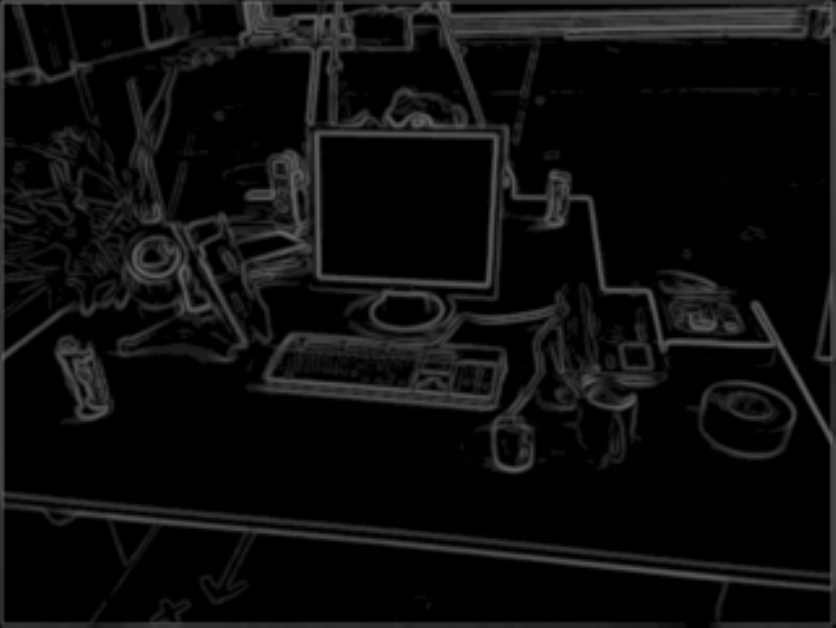}
  \label{fig: Image gradient map}}
  \subfigure[3D edge map.]{
  \includegraphics[width=0.45\columnwidth]{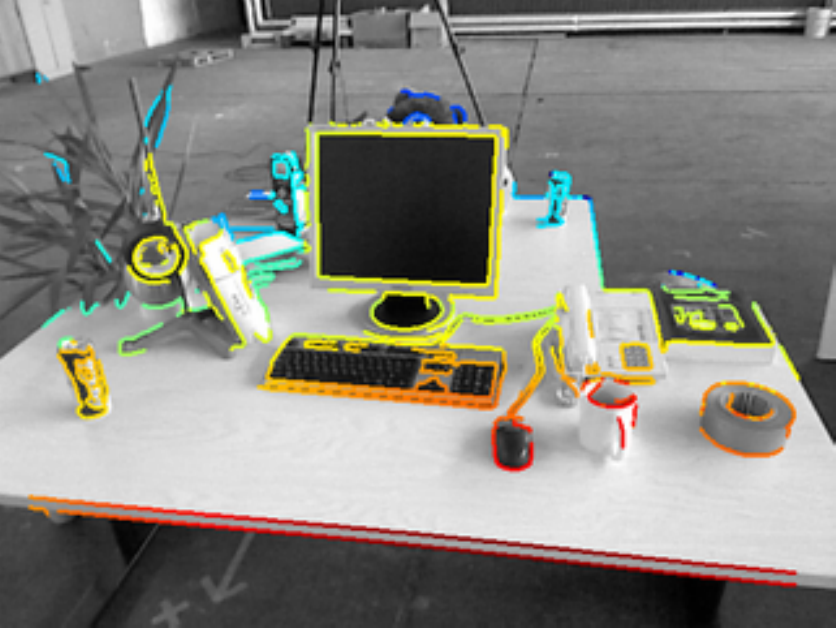}
  \label{fig: semi dense depth map}}
  \caption{Image gradients are calculated in both horizontal and vertical direction at each pixel location. The euclidean norm of each gradient vector is calculated and illustrated in (a) (brighter means bigger while darker means smaller). Canny-edges are obtained by thresholding gradient norms followed by non-maximum suppression. By accessing the depth information of the edge pixels, a 3D edge map (b) is created, in which warm colors mean close points while cold colors represent faraway points. }
  \label{fig: problem definition}
\end{figure}

Let $\mathcal{P}^{\mathcal{F}}=\{\mathbf{p}_{i}^{\mathcal{F}}\}$ be a set of pixel locations in a frame $\mathcal{F}$ defining the 2D edge map. As illustrated in Fig.~\ref{fig: problem definition}, it is obtained by thresholding the norm of the image gradient, which could, in the simplest case, originate from a convolution with Sobel kernels. Let us further assume that the depth value $z_i$ for each pixel in the 2D edge map is available as well. In the preregistered case, they are simply obtained by looking up the corresponding pixel location in the associated depth image. For each pixel, a local patch ($5 \times 5$ pixels) is visited and the smallest depth is selected in the case of a depth discontinuity\footnote{The depths of all pixels in the patch are sorted and clustered based on a simple Gaussian noise assumption. If there exists a cluster center that is closer to the camera, the depth value of the current pixel will be replaced by the depth of that center. This circumvents resolution loss and elimination of fine depth texture.}. This operation ensures that we always retrieve the foreground pixel despite possible misalignments caused by extrinsic calibration errors (between the depth camera and the RGB camera) or asynchronous measurements (RGB and depth) under motion. An exemplary result is given in Fig.~\ref{fig: semi dense depth map}. We furthermore assume that both the RGB and the depth camera are fully calibrated (intrinsically and extrinsically). Thus, we have accurate knowledge about a world-to-camera transformation function $\pi(\lambda \mathbf{f}_{i})=\mathbf{p}_{i}$ projecting any point along the ray defined by a unit vector $\mathbf{f}_i$ onto the image location $\mathbf{p}_i$. The inverse transformation $\pi^{-1}(\mathbf{p}_{i})=\mathbf{f}_{i}$ which transforms points in the image plane into unit direction vectors located on the unit sphere around the center of the camera is also known. If the RGB image and the depth map are already registered, the extrinsic parameters can be omitted. Our discussion will be based on this assumption from now on.

Consider the 3D edge map (defined in the reference frame $\mathcal{F}_\textup{ref}$) as a curve in 3D, and its projection into the current frame $\mathcal{F}_{k}$ as a curve in 2D. The goal of the alignment step is to retrieve the pose at the current frame $\mathcal{F}_{k}$ (namely its position $\mathbf{t}$ and orientation $\mathbf{R}$) such that the projected 2D curve aligns well with the 2D edge map $\mathcal{P}^{\mathcal{F}_{k}}$ extracted in the current frame $\mathcal{F}_{k}$. Note that---due to perspective transformations---this is of course not a one-to-one correspondence problem. Also note that we parametrize our curves by a set of points originating from pixels in a reference image. While there are alternative parameterizations (e.g. splines), the objective function outlined in this work will remain applicable to any parametrization of the structure.
\subsection{ICP-based motion estimation}
\label{subsec: icp based motion estimation}

The problem can be formulated as follows. Let
\begin{equation}
\mathcal{S}^{{\mathcal{F}}_\textup{ref}}
= \bigg\{ \mathbf{s}_{i}^{\mathcal{F}_\textup{ref}} \bigg\}
= \bigg\{ d_{i}^{\mathcal{F}_\textup{ref}}\pi^{-1}\big(\mathbf{p}_{i}^{\mathcal{F}_\textup{ref}}\big) \bigg\}
\label{eq:sdmap}
\end{equation}
denote the 3D edge map in reference frame $\mathcal{F}_\textup{ref}$, where $d_{i} = \frac{z_{i}}{\mathbf{f}_{i,3}}$ denotes the distance of point $\mathbf{s}_{i}$ to the optical center. Its projection onto the current frame $\mathcal{F}_{k}$ results in a number of 2D points%
\begin{equation}
\mathcal{O}^{\mathcal{F}_{k}}
= \bigg\{ \mathbf{o}_{i}^{\mathcal{F}_{k}} \bigg\}
= \bigg\{ \pi \Big( \mathbf{R}^\textup{T} \big( \mathbf{s}_{i}^{\mathcal{F}_\textup{ref}} - \mathbf{t} \big) \Big) \bigg\}.
\end{equation}
We define
\begin{equation}
n(\mathbf{o}_{i}^{\mathcal{F}_{k}}) = \underset{\mathbf{p}_{j}^{\mathcal{F}_{k}} \in \mathcal{P}^{\mathcal{F}_{k}}}{\operatorname{argmin}} \| \mathbf{p}_{j}^{\mathcal{F}_{k}} - \mathbf{o}_{i}^{\mathcal{F}_{k}} \|
\label{eq:nn}
\end{equation}
to be a function that returns the nearest neighbour of $\mathbf{o}_{i}^{\mathcal{F}_{k}}$ in $\mathcal{P}^{\mathcal{F}_{k}}$ under the Euclidean distance metric. The overall objective of the registration is to find
\begin{equation}
\hat{\mathbold{\theta}}=\underset{\mathbold{\theta}}{\operatorname{argmin}} \sum_{i=1}^{N} \|  \mathbf{o}_{i}^{\mathcal{F}_{k}} - n(\mathbf{o}_{i}^{\mathcal{F}_{k}})\|^{2},
\label{eq:sol1}
\end{equation}
where $\mathbold{\theta}:=[t_x,t_y,t_z,c_1,c_2,c_3]^\textup{T}$ represents the parameter vector that defines the pose of the camera. $c_1,c_2,c_3$ are Cayley parameters~\cite{cayleyparameter} for orientation $\mathbf{R}$\footnote{Note that the orientation is always optimized as a change with respect to the previous orientation in the reference frame. The chosen Cayley parametrization therefore is proportional to the local tangential space at the location of the previous quaternion orientation and, therefore a viable parameter space for local optimization of the camera pose.}, and $\mathbf{t} = [t_x,t_y,t_z]^\textup{T}$. The above objective is of the same form as the classical ICP problem, which alternates between finding approximate nearest neighbours and registering those putative correspondences, except that in the present case, the correspondences are between 2D and 3D entities. A very similar objective function has been already exploited by \cite{BMVC2015_100} for robust 3D-2D edge alignment in a hypothesis-and-test scheme. It proceeds by iterative sparse sampling and closed-form registration of approximate nearest neighbours.

\subsection{Euclidean Distance Fields}
\label{subsec:Distance fields}
As outlined in \cite{BMVC2015_100}, the repetitive explicit search of nearest neighbours is too slow even in the case of robust sparse sampling. This is due to the fact that all distances need to be computed in order to rank the hypotheses, and this would again require an exhaustive nearest neighbour search. This is where distance transforms come into play. The explicit location of a nearest neighbour does not necessarily matter when evaluating the optimization objective function (Eq.~\ref{eq:sol1}), the distance alone may already be sufficient. Therefore, we can pre-process the 2D edge map in the current frame and derive an auxiliary image in which the value at every pixel simply denotes the Euclidean distance to the nearest point in the original 2D edge map. Euclidean distance fields can be computed very efficiently using region growing techniques. Chebychev distance is an alternative when faster performance is required. For further information, the interested reader is referred to~\cite{fabbri20082d}.

Let us define $d(\mathbf{o}_{i}^{\mathcal{F}_{k}})$ as the function that retrieves the distance to the nearest neighbour by simply looking up the value at $\mathbf{o}_{i}^{\mathcal{F}_{k}}$ inside the chosen distance field. The optimization objective (Eq.~\ref{eq:sol1}) can now easily be rewritten as
\begin{equation}
\hat{\mathbold{\theta}}=\underset{\mathbold{\theta}}{\operatorname{argmin}} \sum_{i=1}^{N}  d(\mathbf{o}_{i}^{\mathcal{F}_{k}})^{2}.
\label{eq:sol2}
\end{equation}
Methods based on Eq.~\ref{eq:sol2} cannot provide satisfactory performance in terms of efficiency, accuracy and robustness because of the following problems:

\begin{itemize}
\item As pointed out by Kuse \etal ~\cite{kuse2016robust}, the objective function (Eq.~\ref{eq:sol2}) is not continuous due to the spatial discretization of the distance field. This problem is bypassed by for example sampling the distances using bi-linear interpolation. However, even with bi-linear interpolation, the distance remains only a piece-wise smooth (i.e. bi-linear) function, as the parametrization changes depending on which interpolation points are chosen. Kuse \etal ~\cite{kuse2016robust} propose to solve this problem by employing the sub-gradient method, which succeeds in the presence of non-differentiable kinks in the energy function. Rather than employing the more popular Gauss-Newton or Levenberg-Marquardt method, they also rely on a less efficient steepest descent paradigm. While solving the problem, the bi-linear interpolation and the sub-gradient computation increase the computational burden, and the steepest descent method requires more iterations as the inter-frame disparity becomes larger. To guarantee real-time performance, \eg~\cite{kuse2016robust} sacrifies accuracy by working on QVGA resolution. In this work, we advocate the use of nearest neighbour fields, which removes the problem of non-differentiable kinks in the energy function.

\item As explained in \cite{nurutdinova2015towards}, the model-to-data paradigm is affected by a potential bias in the presence of only partial observations. They propose to replace it by a \textit{data-to-model} concept where the summation runs over the measured points in the image. The work parametrizes curves using B-splines, and an additional curve parameter is required for every data point to define the nearest location on the B-spline. This parameter is simply lifted to an additional optimization variable. \cite{nurutdinova2015towards} argues that the \textit{data-to-model} objective is advantageous since it avoids the potentially large biases occurring in the situation of partial occlusions. While the \textit{data-to-model} objective may indeed provide a solution to this problem, it is at the same time a more computational-resource demanding strategy with a vastly blown up parameter space, especially given that the number of automatically extracted pixels along edges can be significantly larger than the number of parameters in a sparse scenario, and one additional parameter for every data point is needed. Furthermore, the lifted optimization problem in \cite{nurutdinova2015towards} depends on reasonably good initial pose estimates that in turn permit the determination of sufficiently close initial values for the curve parameters. In this work, we show how an orientation of the field based on the image gradients effectively counteracts this problem while still enabling the more efficient model-to-data\footnote{Note that a model can refer to either a parametric model (\eg ~a 3D spline curve) or a non-parametric model --- \eg ~a number of 3D edge points --- as in our case.} paradigm.

\item Even ignoring the above two problems, a simple minimization of the L2-norm of the residual distances would fail because it is easily affected by outlier associations. In \cite{BMVC2015_100}, this problem is circumvented by switching to the L1-norm of the residual distances. In this work, we provide a complete analysis of the statistical properties of the residuals, from which we derive an iterative robust reweighting formulation for 3D-2D curve-registration.
\end{itemize}
\newpage
\section{Approximate Nearest Neighbour Fields}
\label{Sec:Approximate Nearest Neighbour Fields}
\begin{figure}[t]
  \centering
  \includegraphics[width=\columnwidth]{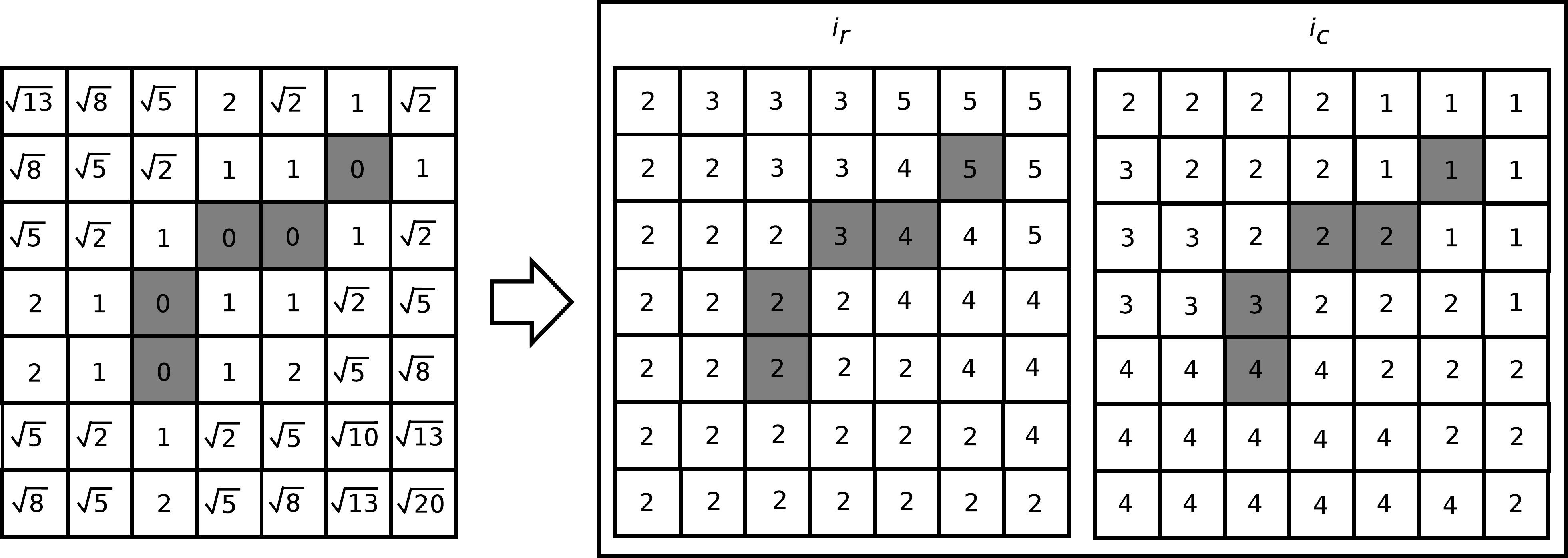}
  \caption{Example of a distance field for a short edge in a 7$\times$7 image, plus the resulting nearest neighbour field. $i_r$ and $i_c$ contain the row and column index of the nearest neighbour, respectively.}
  \label{fig:NNF}
\end{figure}
To solve the first problem, we replace the Euclidean distance fields with approximate nearest neighbour fields. As indicated in Fig.~\ref{fig:NNF}, the nearest neighbour fields consist of two fields indicating the row and the column index of the nearest neighbour, respectively. In other words, the ANNF simply precomputes the expression $n(\mathbf{o}_i)$ in our optimization objective (Eq.~\ref{eq:sol1}) for every possible pixel location in the image. Using ANNFs enables us to fix the nearest neighbours during the Jacobian computation, thus removing the problems of discontinuities or non-smoothness during energy minimization. At the same time, the residual evaluation remains interpolation-free, which relieves the computational burden.

From an implementation point of view, it is important to note that the computation of the nearest neighbour field is equally fast as the derivation of the distance field. The reason lies in the concept of distance field extraction methods \cite{fabbri2008,felzenszwalb12}, which typically perform some sort of region growing, all while keeping track of nearest neighbours in the seed region when propagating pixels. Whether we extract a distance field or a nearest neighbour field is merely a question of which information is retained from the computation.
\subsection{Point-to-Tangent Registration}
\begin{figure}[b]
  \centering
  \includegraphics[width=0.6\columnwidth]{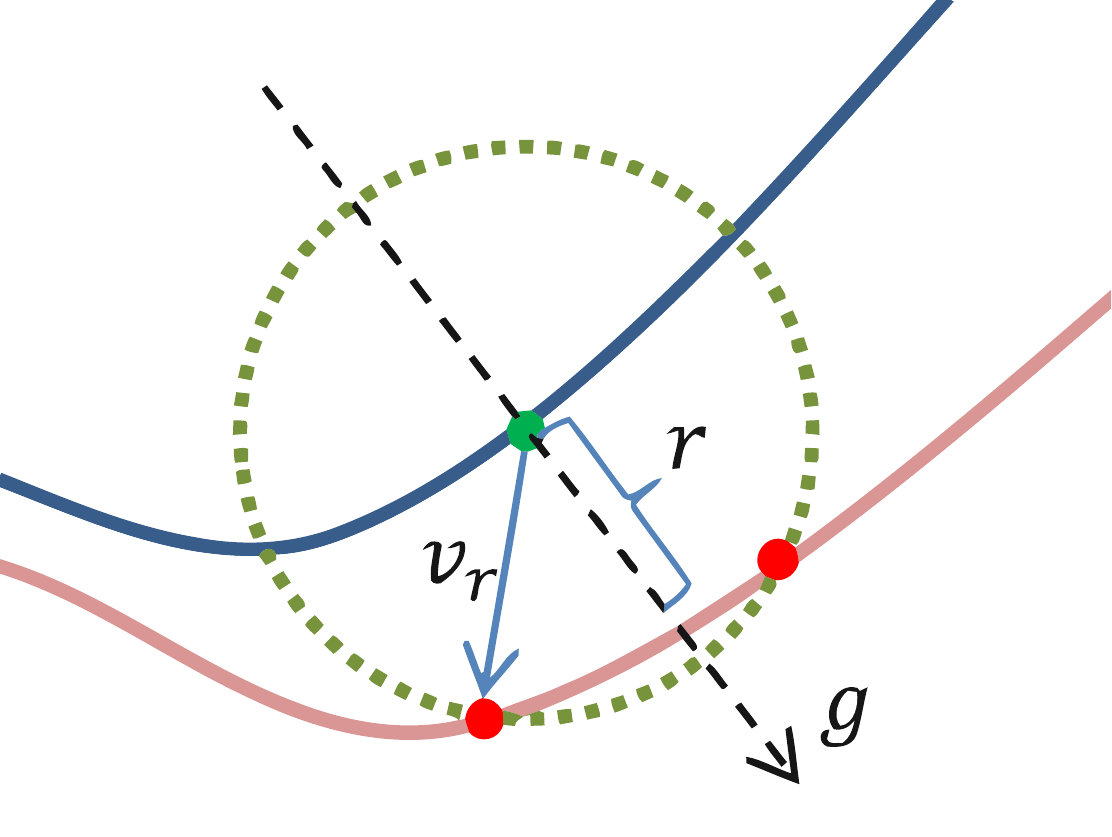}
  \caption{Illustration of the point-to-tangent distance. The projected distance $r$ is finally calculated by projecting $\mathbf{v}_r$ onto the direction of the local gradient $\mathbf{g}$.}
  \label{fig: point to tangential line distance}
\end{figure}
The ICP algorithm and its variants commonly apply two distance metrics in the registration of 3D point cloud data --- the point-to-point distance~\cite{champleboux1992accurate} and the point-to-plane distance~\cite{chen1992object}. ICP using the point-to-plane distance metric is reported to converge faster than the point-to-point version, especially in the so-called \textit{sliding situation}.
In the case of 3D-2D edge alignment, a similar idea to the point-to-plane distance is the point-to-tangent distance. An example is given in Fig.~\ref{fig: point to tangential line distance}, in which the 2D blue curve is the reprojection of the 3D model while the 2D red curve is the data observed in the current frame. Given a point (green) on the blue curve, the coordinate of its closest point (one of the red points) is returned by the ANNF. The point-to-point residual vector is denoted by $\mathbf{v}_r$ and the point-to-tangent distance is obtained by projecting $\mathbf{v}_r$ to the local gradient direction at the green point. Note that the local gradient $\mathbf{g}$ is originally calculated at the corresponding model point in the reference frame. In other words, the gradient $\mathbf{g}$ illustrated in Fig.~\ref{fig: point to tangential line distance} is the warping result of the original gradient vector. This can be done by introducing a hallucinated point, which is along the original gradient vector and with the identical depth as the model point does. Strictly speaking, the gradient direction needs to be recomputed at the beginning of each iteration. However, as we see through our experiments, the gradient direction of each model point can be assumed constant if there is no big rotation between the reference frame and the current frame. Note that the image gradient information is already computed during the edge detection process, thus it does not involve any additional burden of computation. Also note that for EDFs based methods, only $\Vert \mathbf{v}_r \Vert$ is available. Thus, the point-to-tangent distance is not applicable in EDFs.

\subsection{ANNF based Registration}
\label{subsec: ANNF based registration}
Using the ANNF, the function $n(\mathbf{o}_{i}^{\mathcal{F}_{k}})$ from Eq.~\ref{eq:nn} now boils down to a trivial look-up followed by a projection onto the local gradient direction. This enables us to go back to objective (Eq.~\ref{eq:sol1}), and we attempt a solution via efficient Gauss-Newton updates. Let us define the point-to-tangent residuals
\begin{equation}
  \mathbf{r}=\left[ \begin{matrix}
      \mathbf{g}(\mathbf{p}_1^{\mathcal{F}_{\text{ref}}})^\textup{T}\left( \mathbf{o}_1^{\mathcal{F}_k} - n(\mathbf{o}_1^{\mathcal{F}_k}) \right)\\
      \ldots \\
      \mathbf{g}(\mathbf{p}_N^{\mathcal{F}_{\text{ref}}})^\textup{T}\left( \mathbf{o}_N^{\mathcal{F}_k} - n(\mathbf{o}_N^{\mathcal{F}_k}) \right)
  \end{matrix} \right]_{N \times 1}.
\label{eq:reproj_error2}
\end{equation}
By applying Eq.~\ref{eq:reproj_error2} in Eq.~\ref{eq:sol1}, our optimization objective can be reformulated as
\begin{equation}
  \hat{\mathbold{\theta}} = \underset{\mathbold{\theta}}{\operatorname{argmin}} \| \mathbf{r} \|^{2}.
\label{eq:sol3}
\end{equation}

Supposing that $\mathbf{r}$ were a linear expression of $\mathbold{\theta}$, it is clear that solving Eq.~\ref{eq:sol3} would be equivalent to solving $\mathbf{r}(\mathbold{\theta})=\mathbf{0}$. The idea of Gauss-Newton updates (or iterative least squares) consists of iteratively performing a first-order linearization of $\mathbf{r}$ about the current value of $\mathbold{\theta}$, and then each time improving the latter by solving the resulting linear least squares problem. The linear problem to solve in each iteration therefore is given by
\begin{equation}
  \mathbf{r}(\mathbold{\theta}_{i})+\left. \frac{\partial \mathbf{r}(\mathbold{\theta})}{\partial \mathbold{\theta}}\right|_{\mathbold{\theta}=\mathbold{\theta_i}} \mathbold{\Delta} = \mathbf{0},
  \label{eq:itlinsol}
\end{equation}
and, using $\mathbf{J}=\left. \frac{\partial \mathbf{r}(\mathbold{\theta})}{\partial \mathbold{\theta}}\right|_{\mathbold{\theta}=\mathbold{\theta_i}}$, its solution is given by
\begin{equation}
  \mathbold{\Delta} = - (\mathbf{J}^\textup{T}\mathbf{J})^{-1} \mathbf{J}^\textup{T}\mathbf{r}(\mathbold{\theta}_{i}).
\end{equation}
The motion vector is finally updated as $\mathbold{\theta}_{i+1} = \mathbold{\theta}_{i}+\mathbold{\Delta}$.

While evaluating the Jacobian $\mathbf{J}$ in each iteration, the closest points simply remain fixed. This simplification is based on the fact that typically $n(\mathbf{o}_i(\boldsymbol\theta)) = n(\mathbf{o}_i(\boldsymbol{\theta}+\delta{\boldsymbol{\theta}}))$ if $\delta{\boldsymbol{\theta}}$ is a small increment. Furthermore, the equality may not hold when $\mathbf{o}_i$ locates exactly at the border of two pixels. This may lead to gross errors in the Jacobian evaluation, which is why we simply fix the nearest neighbour. The Jacobian $\mathbf{J}$ simply becomes
\begin{equation}
  \mathbf{J} = \left[ \begin{matrix}
    \Big( \frac{\partial \big( \mathbf{g}(\mathbf{p}_1^{\mathcal{F}_{\text{ref}}} )^\textup{T}\mathbf{o}_1^{\mathcal{F}_k}\big)}{\partial \mathbold{\theta}} \Big)^\textup{T} &
    \ldots &
    \Big( \frac{\partial \big( \mathbf{g}(\mathbf{p}_N^{\mathcal{F}_{\text{ref}}} )^\textup{T}\mathbf{o}_N^{\mathcal{F}_k}\big)}{\partial \mathbold{\theta}} \Big)^\textup{T}
  \end{matrix} \right]_{\mathbold{\theta}=\mathbold{\theta_i}}^\textup{T}.
  \label{eq:newjacobian}
\end{equation}
Details on the analytical form of the Jacobian are given in Appendix.~\ref{subsec: appendix derivation of J}.

\section{Oriented Nearest Neighbour Fields}
\label{sec:oriented nearest neighbour fields}
This section explains the idea behind oriented nearest neighbour fields (ONNF) and how they help to improve the performance of model-to-data based projective registration of non-parametric curves. We start by giving a clear definition of the field orientation for distance fields, then show how this design is easily employed to nearest neighbour fields. Finally, a sequence of modifications to this concept is introduced, which gradually improve the accuracy and efficiency of the registration process.

\subsection{Field Orientation}
One of the core contributions of this paper is on orienting the nearest neighbour fields. However, special care is needed to define what \textit{orientation} in the present case means. We explain the concept with distance fields. The most common type of oriented distance field in the 3D computer vision literature is a truncated signed distance field for dense 3D surface reconstruction \cite{newcombe2011kinectfusion,bylow13,steinbruecker13}. Given the fact that the world is always observed from a certain perspective, it makes sense to define the \textit{front} and \textit{back} of a continuous reconstructed surface, which in turn defines the sign of the distances in the field (positive = in front of the surface, negative = behind the surface). In the context of curves in the image, the equivalent would be to define the \textit{inside} and \textit{outside} of contours. This representation, however, would only be unique for a single, closed contour in the image.

A more flexible orientation can be achieved by considering the gradient inclination along the edge. %
The registration bias due to partial occlusions in the model-to-data approach, as pointed out by \cite{nurutdinova2015towards}, could easily be detected or even avoided by considering the ``disparity'' between the reprojected gradient vector inclinations and the gradient vector inclinations of the nearest neighbours in the data. We therefore move to oriented distance fields for curves in the image, where the orientation depends on the actual gradient vector inclination.

\begin{figure}[h]
  \centering
  \subfigure[Discretization bins for gradient vector inclination.]{
  	\includegraphics[width=\columnwidth]{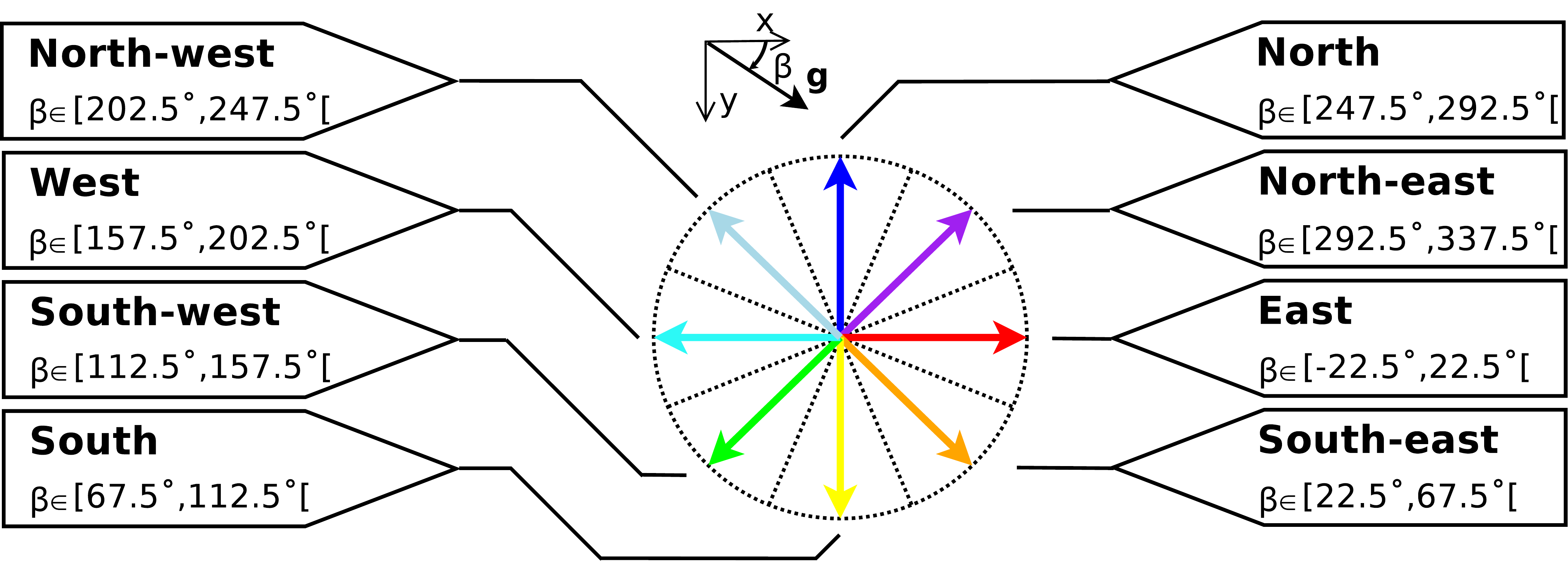}
	\label{fig:orientationBins}
  }
  \subfigure[Example oriented distance fields.]{
  	\includegraphics[width=\columnwidth]{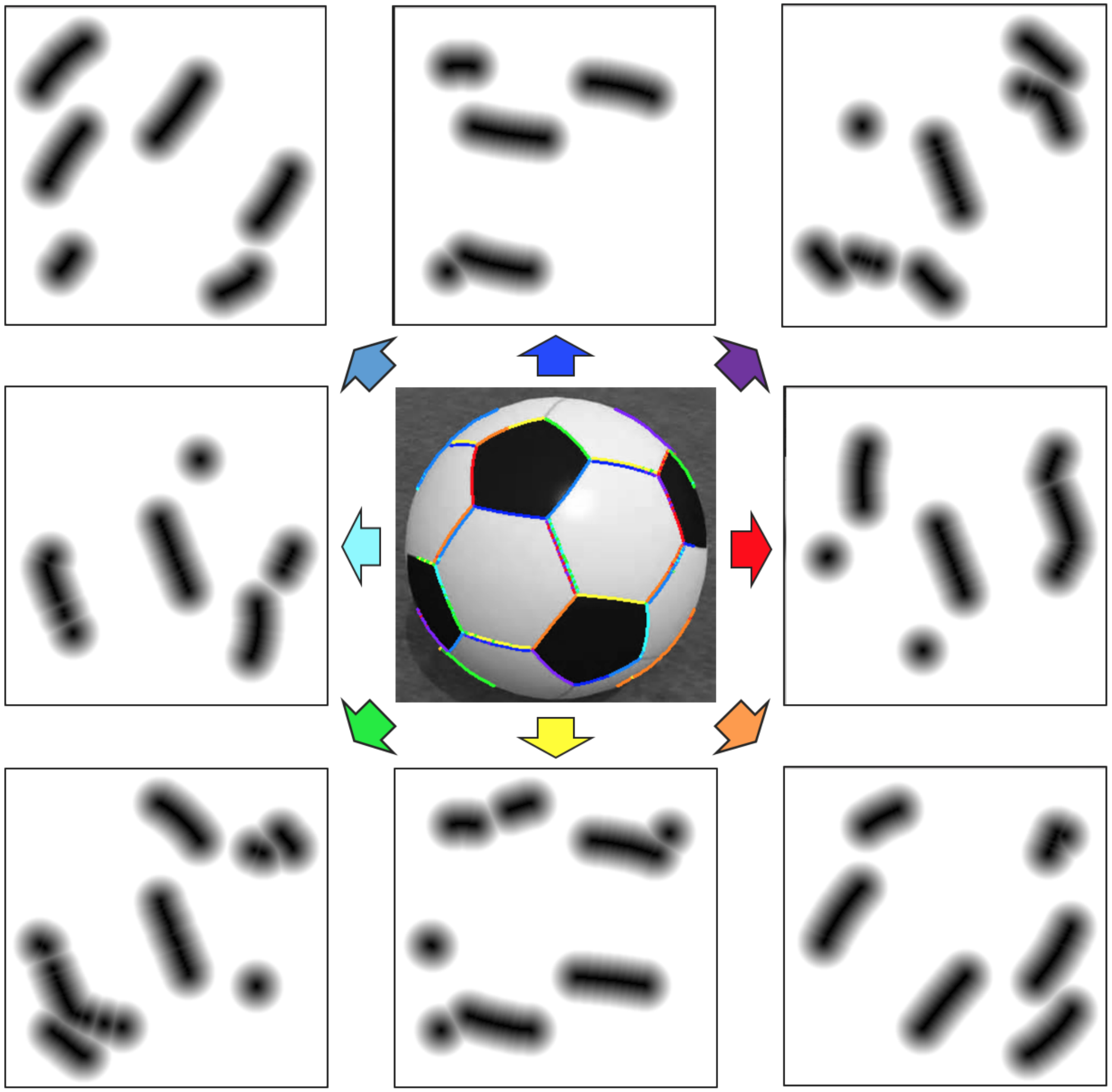}
	\label{fig:orientation}
  }
  \label{fig:orientedDFs}
  \caption{(a) Orientation bins chosen for the discretisation of the gradient vector inclination (8 bins of 45$^{\circ}$ width). (b) Example oriented distance fields for edges extracted from an image of a football. Distinct edge segments are associated to only one of the eight distance fields depending on the local gradient inclination and the corresponding bin.}
\end{figure}

The idea is straightforward. The distance field is split up into multiple distance fields following a quantisation of the gradient vector inclination. The gradient quantisation adopted in this paper is indicated in Fig.~\ref{fig:orientationBins}. It consists of dividing the set of possible gradient vector inclinations into eight equally wide intervals, each one spanning an angle of 45$^{\circ}$. Based on this quantisation table and the local image gradient vector inclination, every pixel along an edge can be associated to exactly one of eight distance fields. We finally obtain a seed region in each one of eight distinct distance fields, and can grow each one of them individually, thus resulting in eight distance fields with exclusively positive numbers (cf. Fig.~\ref{fig:orientation}). Upon registration of a 3D curve, we only need to transform the local gradient of the 3D curve in order to identify the distance field from which the distance to the nearest neighbour of a particular image point has to be retrieved. This formulation has the clear advantage of being less affected by ambiguous associations arising from nearby edges: the distance to the region of attraction of neighbouring edges in the oriented distance field is much larger than in the non-oriented case where all edges appear in the same distance field. In consequence, oriented distance fields also provoke an enlargement of the convergence basin during registration.

Note that the usage of oriented distance fields does not additionally involve any substantial computation load. First, the image gradient information is already computed by the edge extraction algorithm. Second, since the complexity of extrapolating a distance field depends primarily on the number of edge points in the seed region, computing the oriented distance fields is similarly fast as computing the non-oriented one. Furthermore, the orientation makes it very easy to parallelise the distance field computation, we merely have to associate one core to each bin of the discretisation.
\subsection{ONNF based Registration}
As shown in Section~\ref{Sec:Approximate Nearest Neighbour Fields}, distance fields can be seamlessly replaced with nearest neighbour fields. Thus, the concept of the field orientation is also able to be employed to nearest neighbour field, which leads to oriented nearest neighbour fields.

Let us define the nearest neighbour in the oriented nearest neighbour field to be
\begin{equation}
\eta_{\mathcal{M}_{\mathcal{G}\left(\mathbf{o}_{i}^{\mathcal{F}_{k}}\right)}}\left( \mathbf{o}_{i}^{\mathcal{F}_{k}} \right) = \underset{\mathbf{m}_{j}\in\mathcal{M}_{\mathcal{G}\left(\mathbf{o}_{i}^{\mathcal{F}_{k}} \right)}}{\operatorname{argmin}} \| \mathbf{m}_{j}^{\mathcal{F}_{k}} - \mathbf{o}_{i}^{\mathcal{F}_{k}} \|_{2}, 
\end{equation}
with $\mathcal{G}\left(\mathbf{o}_{i}^{\mathcal{F}_{k}}\right)$ taking the gradient at the model point corresponding to $\mathbf{o}_{i}^{\mathcal{F}_{k}}$ and the current camera pose to find the index of the relevant orientation bin (i.e. the index of the relevant nearest neighbour field), and $\mathcal{M}_{\mathcal{G}\left(\mathbf{o}_{i}^{\mathcal{F}_{k}}\right)}$ representing the subset of edge pixels that have fallen into this bin. Similar to what has been proposed in~\ref{subsec: ANNF based registration}, the residual vectors are projected onto the local gradient direction. Since we are already working in an oriented nearest neighbour filed, this gradient direction is simply approximated by the centre of the corresponding orientation bin, denoted $\mathbf{e}_{\mathcal{G}\left(\mathbf{o}_{i}^{\mathcal{F}_{k}}\right)}$ (as in Fig.~\ref{fig:orientationBins}, the possible $\mathbf{e}_{\mathcal{G}\left(\mathbf{o}_{i}^{\mathcal{F}_{k}}\right)}$ are given by the coloured vectors, normalised to one). The residuals can finally be defined as 
\begin{equation}
  \mathbf{r}=\left(\begin{matrix}
  	\mathbf{e}_{\mathcal{G}\left(\mathbf{o}_{1}^{\mathcal{F}_{k}}\right)}^{\text{T}}\big(\mathbf{o}_{1}^{\mathcal{F}_{k}}-\eta_{\mathcal{M}_{\mathcal{G}\left(\mathbf{o}_{1}^{\mathcal{F}_{k}}\right)}}\left( \mathbf{o}_{1}^{\mathcal{F}_{k}} \right)\big) \\
	\vdots \\
	\mathbf{e}_{\mathcal{G}\left(\mathbf{o}_{N}^{\mathcal{F}_{k}}\right)}^{\text{T}}\big(\mathbf{o}_{N}^{\mathcal{F}_{k}}-\eta_{\mathcal{M}_{\mathcal{G}\left(\mathbf{o}_{N}^{\mathcal{F}_{k}}\right)}}\left( \mathbf{o}_{N}^{\mathcal{F}_{k}} \right)\big)
  \end{matrix}\right),
\end{equation}
and the resulting Jacobian becomes
\begin{equation}
  \mathbf{J}= \left[\begin{matrix}
  	\left(  \mathbf{e}_{\mathcal{G}\left(\mathbf{o}_{1}^{\mathcal{F}_{k}}\right)}^{\text{T}}  \frac{\partial   \mathbf{o}_{1}^{\mathcal{F}_{k}}   }{\partial\mathbold{\theta}}\right)^{T} &&
	\ldots &&
	\left(  \mathbf{e}_{\mathcal{G}\left(\mathbf{o}_{N}^{\mathcal{F}_{k}}\right)}^{\text{T}}  \frac{\partial   \mathbf{o}_{N}^{\mathcal{F}_{k}}   }{\partial\mathbold{\theta}}\right)^{T}
  \end{matrix}\right]^{\text{T}}_{\mathbold{\theta}=\mathbold{\theta}_{i}}.
\end{equation}
The derivation of the analytical Jacobian is similar to~\ref{subsec: appendix derivation of J}.
\begin{figure*}[!htb]
  \centering
  \includegraphics[width=0.8\textwidth]{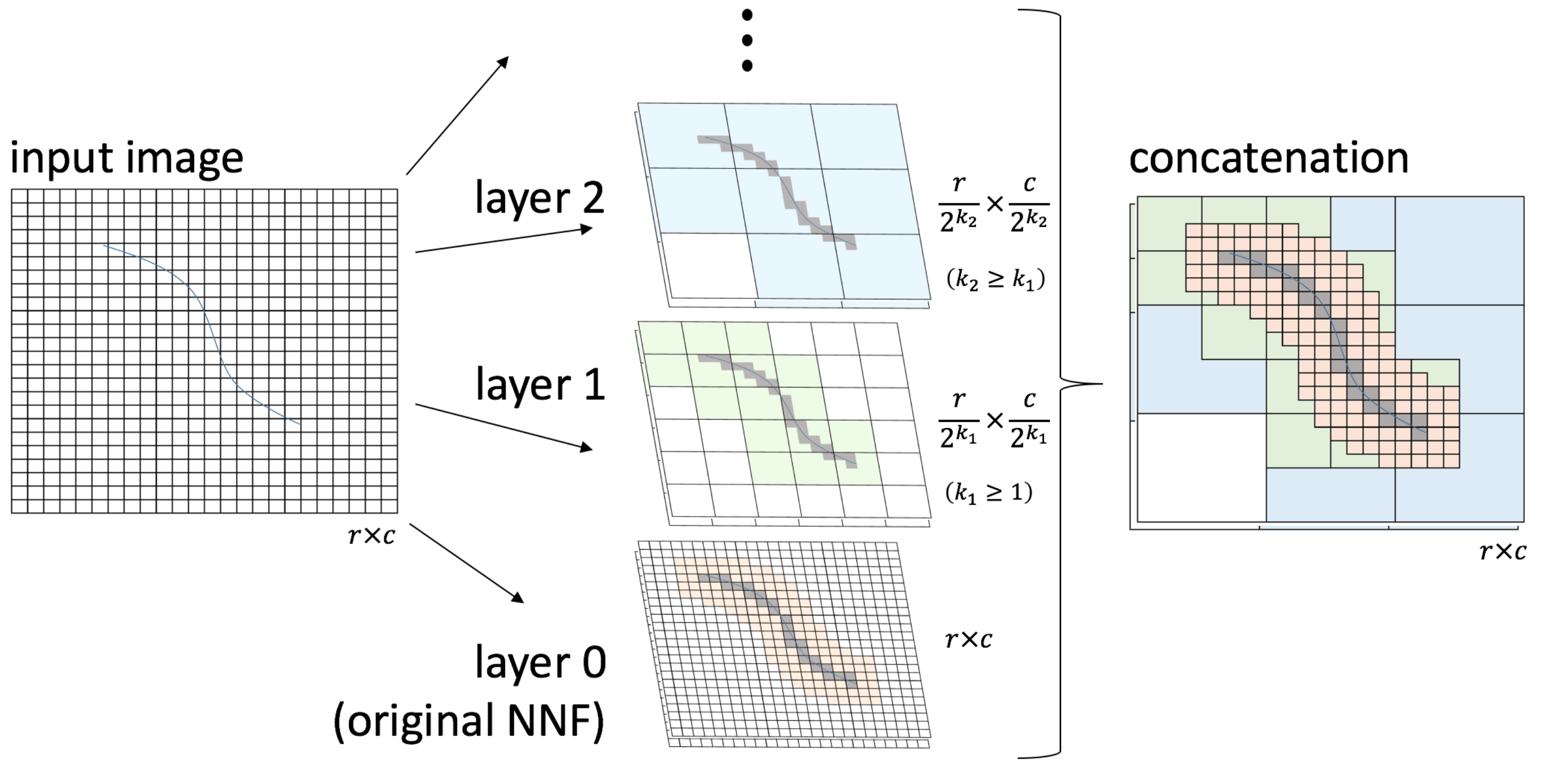}
  \caption{Adaptively Sampled Nearest Neighbour Fields. In practice, the concatenated result is just an $r \times c$ matrix where the connected blue and green regions simply contain identical elements.}
  \label{fig:ASNNF}
\end{figure*}
\subsection{Performance Boost through Adaptive Sampling}

Our final modification consists of moving from standard nearest neighbour fields to adaptively sampled nearest neighbour fields \cite{frisken00}. Nearest neighbours at the original image resolution are only computed within a small neighbourhood of the seed region given by the pixels along edges. With reference to Fig.~\ref{fig:ASNNF}, this corresponds to layer 0. The next step consists of iterating through all edge pixels and keeping track of the closest one to each adjacent location in sub-sampled image grids. Again with reference to Fig.~\ref{fig:ASNNF}, this corresponds to all higher octaves (i.e. layer 1, layer 2, ...). Note that limiting the filling in higher octaves to adjacent grid locations leads to an implicit truncation of the neighbour field. The concluding step then consists of concatenating the layers by copying the nearest neighbours from all layers to the corresponding locations in the concatenated output matrix, starting from the highest one. Values taken from higher octaves are hence simply overwritten if a lower octave contains more fine-grained information. Fig.~\ref{fig:ASNNF} only shows a single nearest neighbour field, but it is clear that the derivation has to be done for each one of the eight orientation bins, possibly through parallel computation.

The adaptively sampled nearest neighbour fields do not involve any loss in accuracy, as the nearest neighbours have maximal resolution within a sufficiently large band around the global minimum. Furthermore, the loss in effective resolution further away from the global minimum does not have a noticeable impact on the ability to bridge even larger disparities. In particular, the fact that the residual vectors are projected onto the direction vector of the corresponding orientation bin causes the approximation error with respect to the exact nearest neighbour to be relatively small. While adaptive sampling is also applicable to distance fields, it would severely complicate the implementation of bi-linear interpolation and hence the definition of continuous residual errors.

A comparison of the properties of all discussed distance transformations is given in Table.~\ref{table:DF property table}, which helps to highlight the advantages of the proposed distance transformations over the classical Euclidean distance field.
\begin{table}[!b]
\centering
\caption{Comparison on the properties of different distance transformations}
\label{table:DF property table}
\begin{tabular}{|c|c|c|c|}
\hline
                                                                                                        & EDF & ANNF & ONNF \\ \hline
\begin{tabular}[c]{@{}c@{}}Free of interpolation\end{tabular}                                           & $\times$    &  $\checkmark$    &  $\checkmark$    \\ \hline
\begin{tabular}[c]{@{}c@{}}Enable point-to-tangent distance\end{tabular}                                & $\times$    &  $\checkmark$    &  $\checkmark$    \\ \hline
\begin{tabular}[c]{@{}c@{}}Enable adaptive sampling\end{tabular}                                        & $\times$    &  $\checkmark$    &  $\checkmark$    \\ \hline
\begin{tabular}[c]{@{}c@{}}Enable registration bias \\ recognition and elimination\end{tabular}         & $\times$    &  $\times$        &  $\checkmark$    \\ \hline
\end{tabular}
\end{table}
\section{Robust Motion Estimation}
\label{sec: robust vo system}

In this section, we discuss how to improve the robustness of the method. A probabilistic formulation is employed in the motion estimation to deal with noise and outliers, which takes the statistical characteristics of the sensor or measurement model into account. Then a simple but effective operation of point culling is introduced, which helps to refine the 3D structure in the reference frame, and thus brings benefits to successive motion estimations. Finally, the whole visual odometry system is outlined.

\subsection{Learning the Probabilistic Sensor Model}

To improve the robustness in the presence of noise and outliers, the motion estimation is formulated as maximizing the posteriori $p(\boldsymbol\theta \vert \mathbf{r})$.  Following the derivation in~\cite{kerl2013robust}, the Maximum A Posteriori (MAP) problem is translated into the weighted least squares minimization problem,
\begin{equation}
\mathbold{\theta} = \argmin_{\mathbold{\theta}} \sum_i \omega(r_i)(r_i(\mathbold{\theta}))^{2}.
\label{eq: weighted least squares}
\end{equation}
The weight is defined as $\omega (r_i) = -\frac{1}{2r_i}\frac{\partial \log p(r_i \vert \mathbold{\theta})}{\partial r_i}$, which is a function of the probabilistic sensor model $p(r_i \vert \mathbold{\theta})$. IRLS is used for solving Eq.~\ref{eq: weighted least squares}. 

The choice of the weight function depends on the statistics of the residual, which is identified in a dedicated experiment. We investigate several of the most widely used robust weight functions including Tukey\footnote{The Tukey-Lambda distribution is used here rather than the Tukey Biweight function. The closed form of the Tukey-Lambda distribution requires to set shape parameter $lambda = 0$, which leads to the Logistic distribution. The derivation of the robust weight function is given in Section ~\ref{App: RWF of Tukey}.}, Cauchy, Huber~\cite{zhang1997parameter} and the T-distribution~\cite{kerl2013robust}. The final choice is based on the result of the model fitting in~\ref{subsec: best configuration}.
\subsection{Point Culling}
Although the probabilistic formulation can deal with noise and outliers, an accurate 3D edge map for each reference frame is still preferred to reduce the risk of an inaccurate registration. Once a new reference frame is created by loading the depth information, the 3D edge map might be not accurate enough because of low-quality depth measurements (e.g. by reflective surfaces) or inaccurate Canny edge detections (e.g. caused by image blur). The successive tracking is possibly affected if the error in the map is not carefully handled. For the sake of computational efficiency, we do not optimize the local map using windowed bundle adjustment as this is commonly done for sparse methods. The number of points used by our method typically lies between $10^{3}$ and $10^{4}$, which is at least one order of magnitude higher than the amount of points used in sparse methods. Therefore, rather than optimizing the inverse depth of such a big number of 3D points, a much more efficient strategy is proposed. All 3D points in the new reference frame are reprojected to the nearest reference frame and those whose geometric residuals are larger than the median of the overall residuals are discarded. We find that this operation significantly improves the accuracy of the motion estimation during our experiments.
\subsection{Visual Odometry System}
\begin{figure}[b!]
  \centering
  \includegraphics[width=0.9\columnwidth]{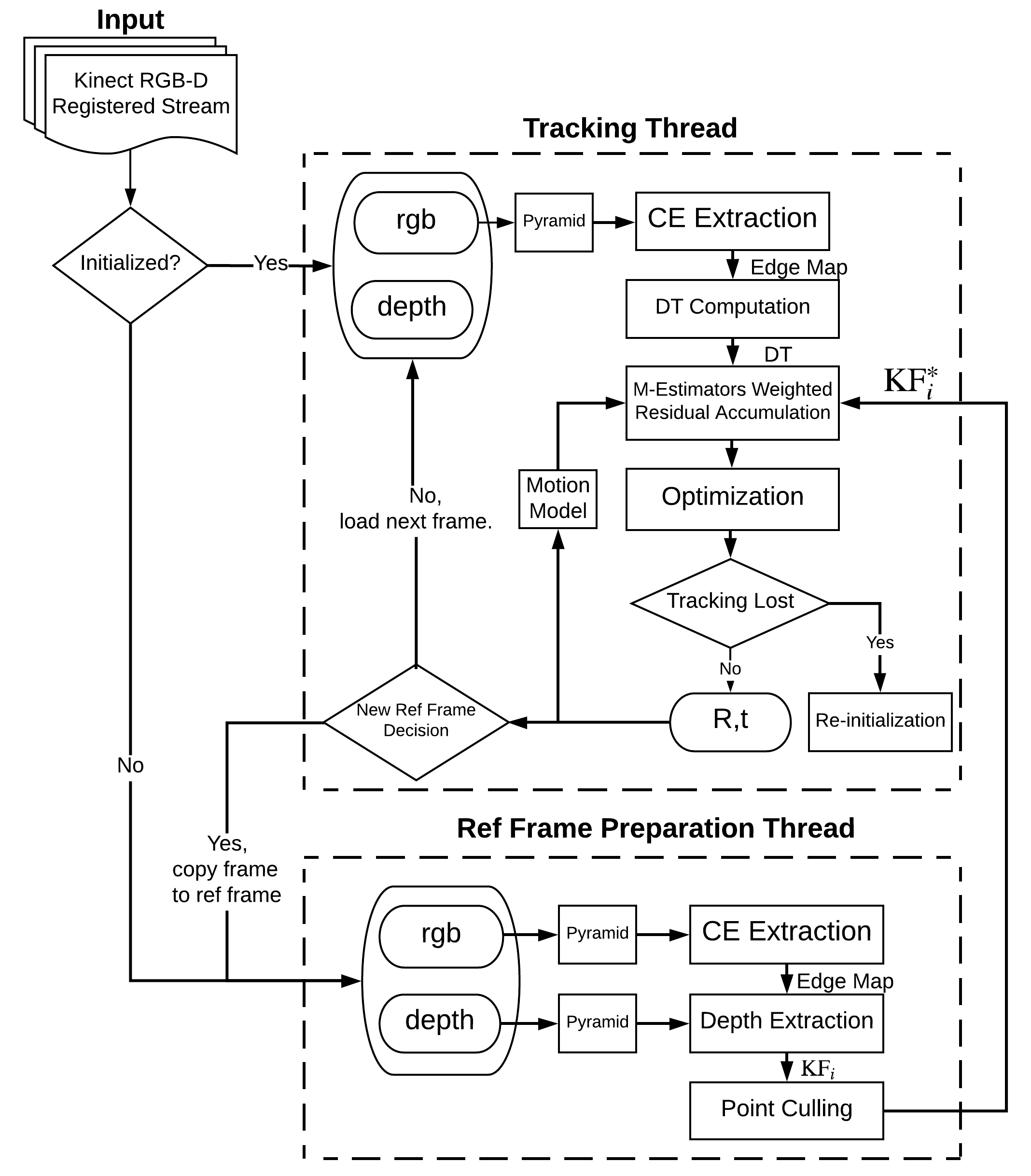}
  \caption{Flowchart of the Canny-VO system. Each independent thread is bordered by a dashed line. CE refers to the Canny edge and DT is the abbreviation of distance transformation, which could be one of EDF, ANNF and ONNF.}
  \label{fig:vo system flowchart}
\end{figure}
Our complete RGB-D visual odometry system is illustrated in Fig.~\ref{fig:vo system flowchart}. There are two threads running in parallel. The tracking thread estimates the pose of the current frame, while the reference frame preparation thread generates new reference frames including the depth initialization. In the tracking thread, only the RGB image of the current frame is used for the Canny edge detection and the subsequent computation of the nearest neighbour field. The objective is constructed and then optimized via the Gauss-Newton method. The reference frame is updated whenever the current frame moves too far away. Our distance criterion here is the median disparity between the edges in the reference frame and the corresponding reprojections in the registered current frame. If this value grows larger than a given threshold, a new reference frame is created by the reference frame preparation thread. The operations of the reference frame preparation thread have been detailed in \ref{Subsec: Problem statement}.

To deal with large displacement, we apply a pyramidal coarse-to-fine scheme as in~\cite{kerl2013robust,engel2013semi}. A three-level (from 0 to 2) image pyramid is created. 
The distance transformation is truncated adjustably according to the applied level.
The registration is performed from the top to the bottom level sequentially. Besides, a motion model is implemented to predict a better starting position for the optimization. This strategy has been widely used in VO and SLAM~\cite{klein2007parallel,kerl2013robust,tanskanen2013live} and improves the robustness by effectively avoiding local minima in the optimization. Instead of assuming a prior distribution for the motion as in~\cite{kerl2013robust}, we follow~\cite{klein2007parallel} and implement a simple decaying velocity model, which effectively improves the convergence speed and the tracking robustness.
\section{Experimental Results}
\label{Sec: Evaluation}
We start with an analysis of the registration bias in the case of partially observed data. We then move over to the optimal parameter choice in our system, which primarily discusses the choice of the robust weight function. Our main experiment compares the quantitative results of trackers that use EDF, ANNF and ONNF, respectively. All variants employ Gauss-Newton method. Two publicly available benchmark datasets~\cite{sturm2012benchmark, handa:etal:ICRA2014} are used for the evaluation. Finally, we provide a challenging RGB-D sequence to qualitatively evaluate the performance of our VO system in a relatively large-scale indoor environment.

Note that the trajectory evaluation results listed in the following tables, including relative pose errors (RPEs) and absolute trajectory errors (ATEs) are given as root-mean-square errors (RMSEs). The units for RPEs are $\deg/s$ and $m/s$ and the ATEs are expressed in $m$. The best result is always highlighted in bold.

\subsection{Handling Registration Bias}
\label{subsec: handling partial occlusion}
\begin{figure}[h!]
\centering
\subfigure[Circle pattern and the partially observed data.]{
\includegraphics[width=0.23\textwidth]{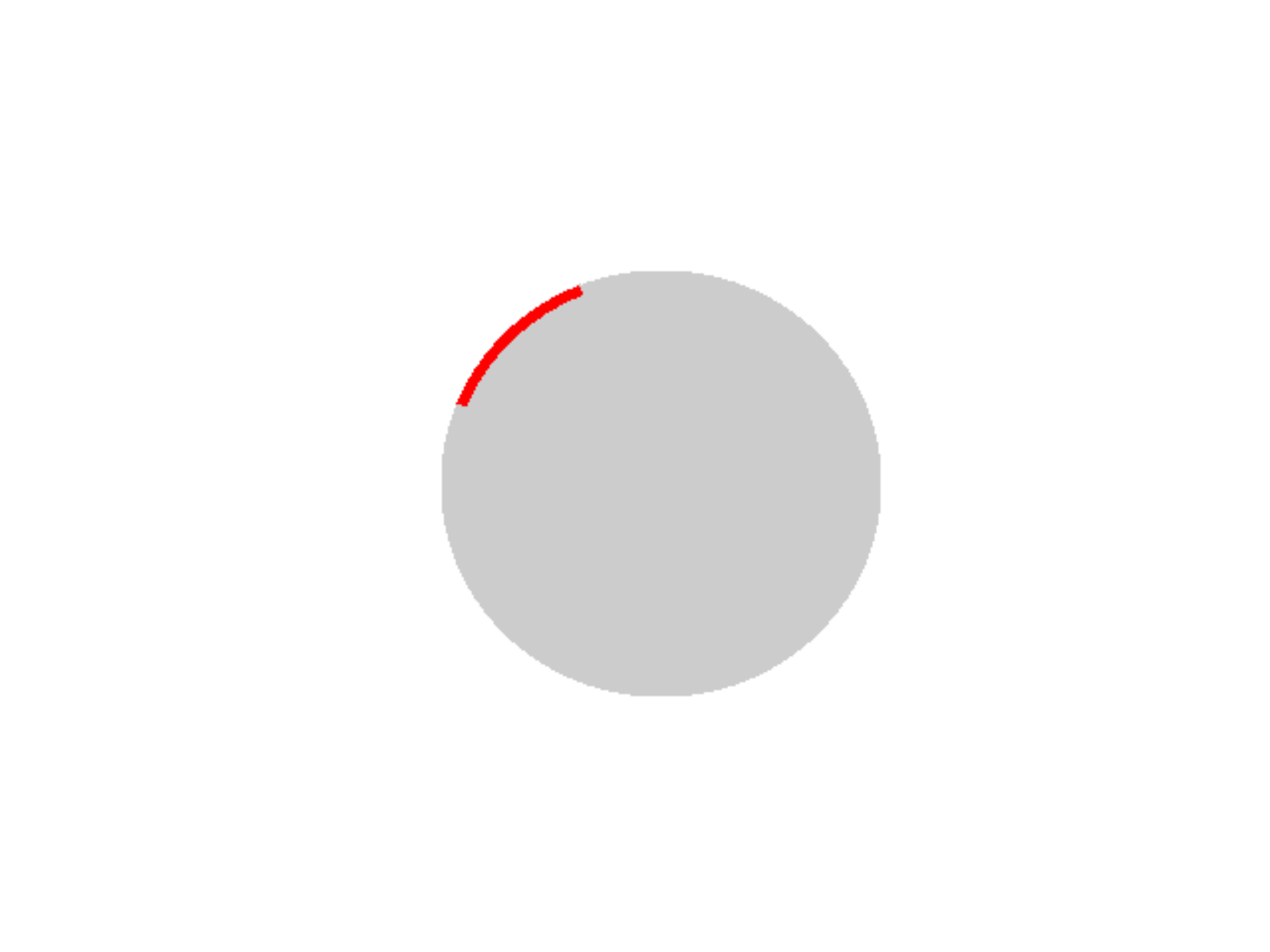}}
\subfigure[Median translation errors of each method for 1000 trials.]{
\includegraphics[width=0.23\textwidth]{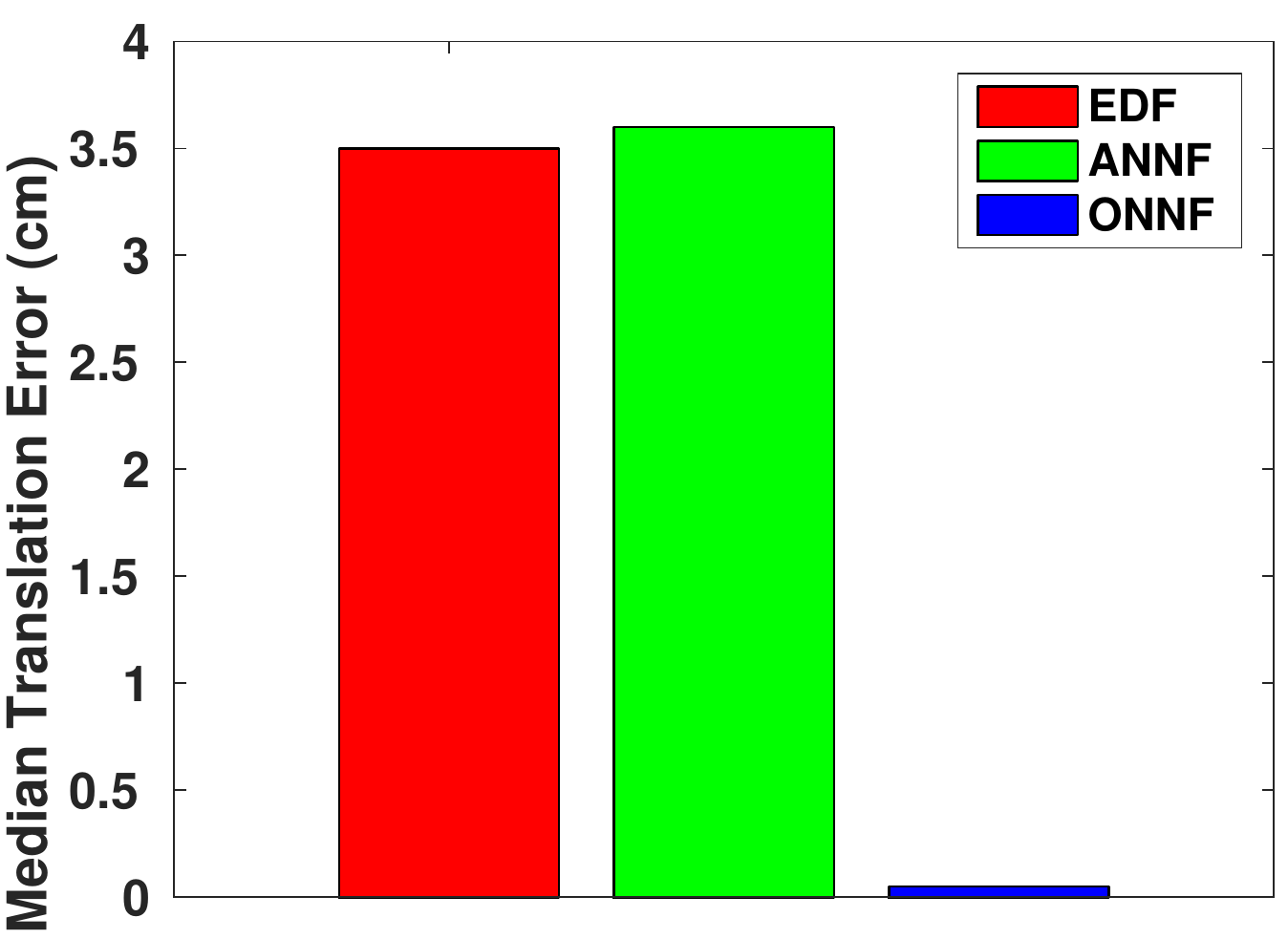}}
\caption{Analysis of registration bias in case of only partially observed data.}
\label{fig: partial occlusion}
\end{figure}

The present section contains an important result of this paper, namely a dedicated experiment on a controlled synthetic sequence to prove the beneficial properties of the presented oriented nearest neighbour fields. We define an artificial circular pattern on the ground plane. The pattern has the size of an A4 sheet of paper. We then assume a downward-looking perspective camera with a focal length of 500.0 and VGA resolution. The height of the camera is chosen such that the pattern covers the entire width of the image. The pose of the reference frame is therefore given by $\mathbf{t}=\left[\begin{matrix} 0 && 0 && 218.75\end{matrix}\right]^{\text{T}}$ and $\mathbf{R}=\operatorname{diag}\left(1,-1,-1\right)$. Once the 3D edge points are extracted, the position of the reference frame is disturbed and re-optimised using either EDF, ANNF or ONNF. To create the cases of partial observation that are very similar to the one introduced in~\cite{nurutdinova2015towards}, only a small continuous part of the circular edge in the image covering $\frac{\pi}{4}$ rad is retained (randomly positioned along the circle). Each method is tested for 1000 times. Note that the tests are not using a robust weight function in order not to hide potential biases in the estimation, which is what we are after. Also note that we do not add any noise to the data as the purpose here is to demonstrate the size of convergence basins, numerical accuracy, and estimation biases. As seen in Fig.~\ref{fig: partial occlusion}, ONNF reports an almost zero bias after optimization, thus clearly demonstrating its superiority in handling partially observed data over the other two methods.

\subsection{Exploring the Optimal Configuration}
\label{subsec: best configuration}

An accurate extraction of Canny edges is key to accurate motion estimation in our method. The quality of the gradient map makes the difference. We therefore investigate Sobel filters with different kernel sizes, and find that a symmetric $5 \times 5$ kernel outperforms a $3 \times 3$ filter and is sufficient for a good estimation. Advance smoothing of the images further helps to improve the edge detection.

\begin{figure*}[h]
\centering
\subfigure[Sensor model of \textsl{freiburg1}.]{
\includegraphics[width=0.3\textwidth]{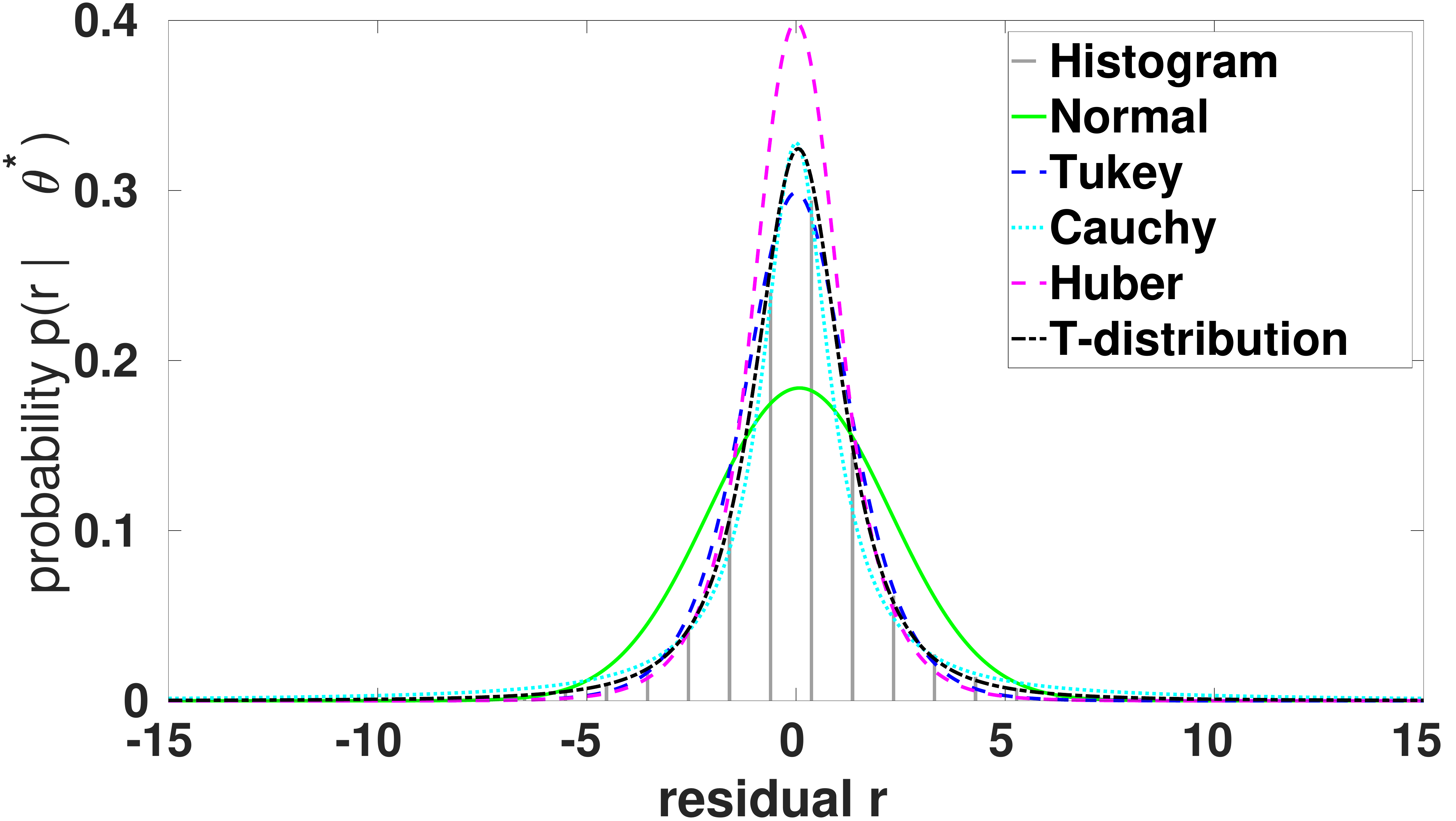}}
\subfigure[Sensor model of \textsl{freiburg2}.]{
\includegraphics[width=0.3\textwidth]{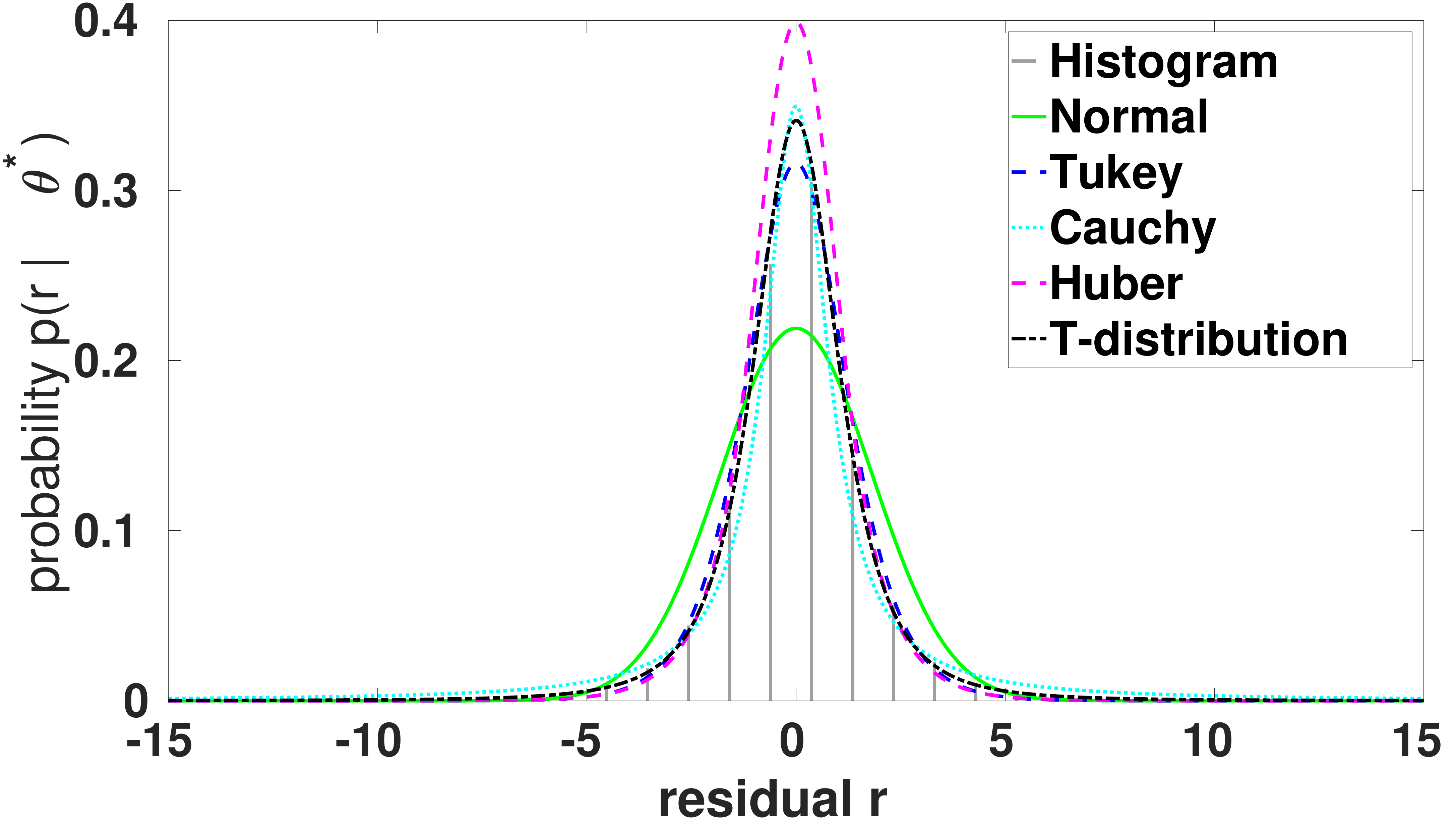}}
\subfigure[Sensor model of \textsl{freiburg3}.]{
\includegraphics[width=0.3\textwidth]{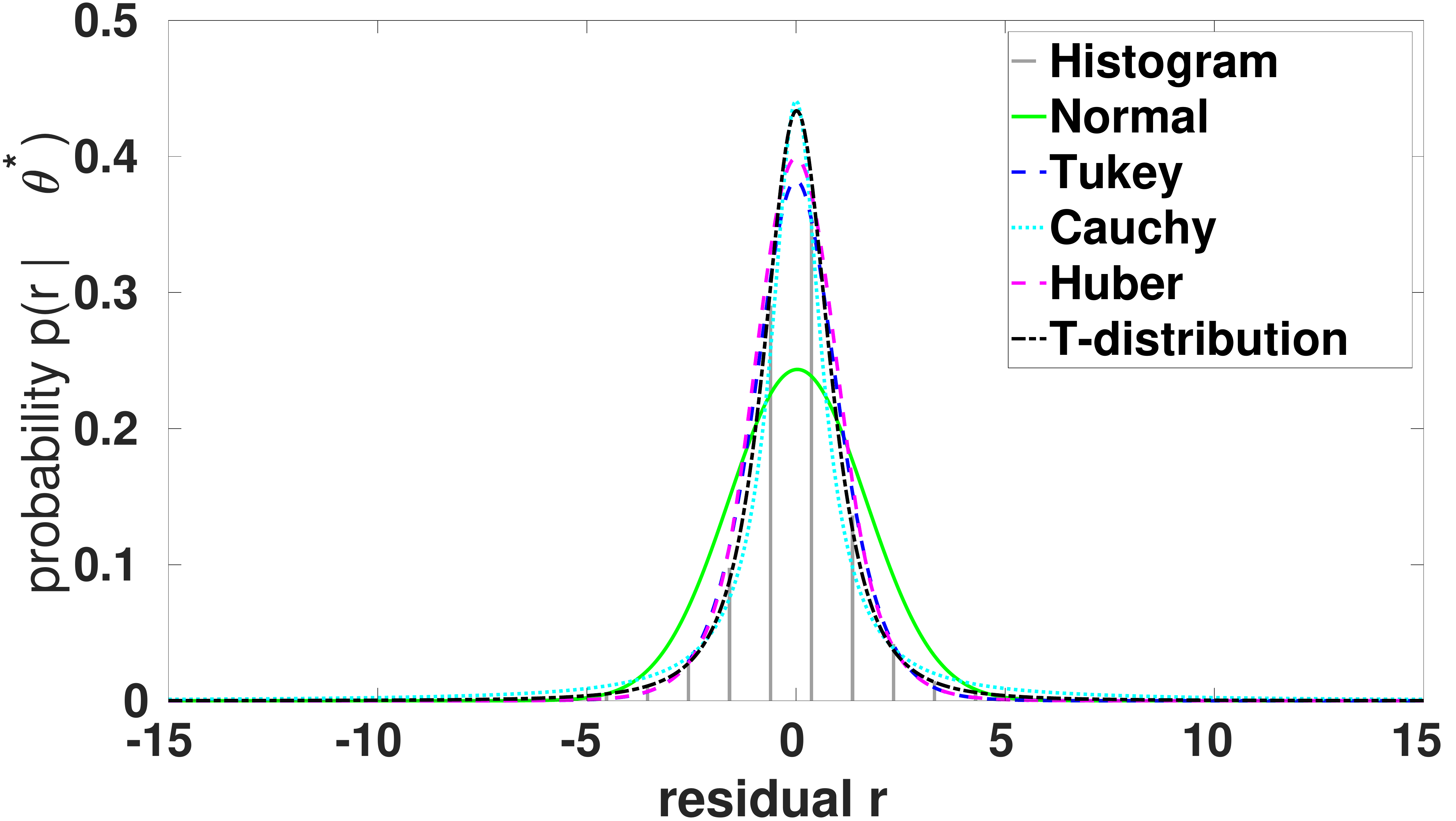}}
\caption{Sensor model $p(r | \boldsymbol{\theta}^{\star})$ is obtained by fitting the histogram with different probabilistic distributions, in which $\boldsymbol{\theta}^{\star}$ denotes the ground truth pose.}
\label{fig: fitted sensor model}
\end{figure*}

To determine the optimal robust weight function, we start by defining reference frames in a sequence by applying the same criterion as in our full pipeline (cf. Fig.~\ref{fig:vo system flowchart}), however using ground truth poses. Residuals are then calculated using the ground truth relative poses between each frame and the nearest reference frame. The residuals are collected over several sequences captured by the same RGB-D camera (\ie~\textit{freiburg 1, freiburg 2, freiburg 3,} respectively), and then summarized in histograms. As an example, all fitting results on the residuals using the ANNF distance metric are illustrated in Fig.~\ref{fig: fitted sensor model}, and the parameters of each model are reported in Table~\ref{table:robust weight function parameters}. The fitting results on the residuals using EDF and ONNF can be obtained in the same way. In general, the T-distribution is the best on fitting the histograms, especially for large residuals. 

\begin{table*}[t]
\centering
\caption{Robust weight functions and their parameters fitted on each sub dataset}
\label{table:robust weight function parameters}
\begin{tabular}{|c|c|c|c|c|}
\hline
Type               & $\omega(r)$                                                           & freiburg1                                               & freiburg2                                               & freiburg3                                               \\ \hline
Tukey          & $\left\{ \begin{array}{ll} \frac{1}{2k^2(e^{r/k} + 1)}, \textup{if} \vert r\vert \leq \epsilon\\
                 \frac{e^{r/k} - 1}{2kr(e^{r/k} + 1)}, \textup{if} \vert r \vert > \epsilon
                 \end{array} \right.$                                                & k = 0.8368 & k = 0.7909 & k = 0.6540                                              \\ \hline
Cauchy         & $\frac{1}{1+(r/k)^2}$                                                 & k = 0.9701 & k = 0.9102 & k = 0.7217                                              \\ \hline
Huber          & $\left\{ \begin{array}{ll} 1, \textup{if} \vert r\vert \leq k\\
                 \frac{k}{\vert r \vert}, \textup{if} \vert r \vert > k
                 \end{array} \right.$                                                  & k = 1.1426                                              & k = 1.1710                                              & k = 1.4425                                              \\ \hline
T-distribution & $\frac{\nu + 1}{\nu + (\frac{r}{\sigma})^2}$  & \begin{tabular}[c]{@{}c@{}} $\nu$ = 2.2875\\ $\sigma$ = 1.1050\end{tabular} & \begin{tabular}[c]{@{}c@{}}$\nu$ = 2.7104\\ $\sigma$ = 1.0682\end{tabular} & \begin{tabular}[c]{@{}c@{}}$\nu$ = 2.4621\\ $\sigma$ = 0.8330\end{tabular} \\ \hline
\end{tabular}
\end{table*}

\subsection{TUM RGB-D benchmark}
\label{subsec:TUM RGB-D benchmark}

The TUM RGB-D dataset collection~\cite{sturm2012benchmark} contains indoor sequences captured using a Microsoft Kinect v.1 sensor with VGA resolution along with ground truth trajectories of the sensor and a set of tools for easily evaluating the quality of the estimated trajectories. We evaluate our methods on almost all the sequences in the dataset except for those in which scenes are beyond the range of the sensor. Our main purpose is to demonstrate the advantage of the proposed ANNF and ONNF over the classical EDF in terms of accuracy and robustness. Since one of state-of-the-art implementations~\cite{kuse2016robust} terminates the optimization on the QVGA resolution, its results are not at the same level. To achieve a fair comparison, we implement our own EDF based tracker which outperforms~\cite{kuse2016robust}. Besides, to comprehensively assess the performance, a sparse feature based solution ORB-SLAM2 (RGB-D version)~\cite{murORB2} is included in the evaluation. Note however that we only use the tracker of~\cite{murORB2} in order to fairly assess pure tracking performance (by setting \textit{mbOnlyTracking}$=$\textit{true}) in the experiment.

\begin{table*}[p]
\centering
\caption{Relative Pose RMSE($\mathbf{R}$:deg/s, $\mathbf{t}$:m/s) of TUM datasets}
\label{table: TUM RPE evaluation}
\begin{tabular}{|c|c|c|c|c|c|c|c|c|}
\hline
                                & \multicolumn{2}{c|}{\begin{tabular}[c]{@{}c@{}}ORB-SLAM2\\ Features\end{tabular}} & \multicolumn{2}{c|}{\begin{tabular}[c]{@{}c@{}}Our Implementation (EDF)\\ Edge Alignment\end{tabular}} & \multicolumn{2}{c|}{\begin{tabular}[c]{@{}c@{}}Our Method (ANNF)\\ Edge Alignment\end{tabular}} & \multicolumn{2}{c|}{\begin{tabular}[c]{@{}c@{}}Our Method (ONNF)\\ Edge Alignment\end{tabular}} \\ \hline
Seq.                            & RMSE(R)                                 & RMSE(t)                                 & RMSE(R)                                            & RMSE(t)                                           & RMSE(R)                                        & RMSE(t)                                        & RMSE(R)                                        & \multicolumn{1}{l|}{RMSE(t)}                  \\ \hline
fr1\_360                        & \textbf{2.593}                          & \textbf{0.065}                          & 15.922                                             & 0.448                                             & 6.450                                          & 0.211                                          & 4.036                                          & 0.121                                         \\ \hline
fr1\_desk                       & 2.393                                   & 0.051                                   & 2.966                                              & 0.073                                             & {\color[HTML]{000000} 6.312}                   & {\color[HTML]{000000} 0.075}                   & \textbf{1.923}                                 & \textbf{0.031}                                \\ \hline
fr1\_desk2                      & \textbf{3.533}                          & \textbf{0.074}                          & 6.109                                              & 0.359                                             & 6.513                                          & 0.156                                          & 5.056                                          & 0.131                                         \\ \hline
fr1\_floor                      & 2.739                                   & 0.038                                   & 1.248                                              & 0.037                                             & 0.880                                          & 0.013                                          & \textbf{0.823}                                 & \textbf{0.010}                                \\ \hline
fr1\_plant                      & 1.837                                   & 0.044                                   & \textbf{1.486}                                     & \textbf{0.028}                                    & {\color[HTML]{000000} 2.864}                   & {\color[HTML]{000000} 0.050}                   & 1.535                                          & 0.036                                         \\ \hline
fr1\_room                       & 2.721                                   & 0.076                                   & 2.762                                              & 0.084                                             & {\color[HTML]{000000} 4.445}                   & {\color[HTML]{000000} 0.223}                   & \textbf{2.003}                                 & \textbf{0.042}                                \\ \hline
fr1\_rpy                        & 2.393                                   & 0.037                                   & 5.430                                              & 0.119                                             & {\color[HTML]{000000} 5.699}                   & {\color[HTML]{000000} 0.063}                   & \textbf{2.245}                                 & \textbf{0.034}                                \\ \hline
fr1\_teddy                      & \textbf{2.061}                          & \textbf{0.062}                          & 4.193                                              & 0.153                                             & {\color[HTML]{000000} 9.248}                   & {\color[HTML]{000000} 0.196}                   & 2.921                                          & 0.123                                         \\ \hline
fr1\_xyz                        & \textbf{0.958}                          & \textbf{0.014}                          & 1.178                                              & 0.024                                             & 1.534                                          & 0.045                                          & 1.127                                          & 0.019                                         \\ \hline
fr2\_360\_hemisphere            & 3.482                                   & 0.280                                   & 7.817                                              & 0.833                                             & 2.751                                          & 0.380                                          & \textbf{1.092}                                 & \textbf{0.108}                                \\ \hline
fr2\_360\_kidnap                & 2.471                                   & 0.174                                   & 3.077                                              & 0.244                                             & 2.151                                          & 0.172                                          & \textbf{1.161}                                 & \textbf{0.084}                                \\ \hline
fr2\_coke                       & 4.845                                   & 0.165                                   & 4.340                                              & 0.102                                             & {\color[HTML]{000000} \textbf{1.179}}          & {\color[HTML]{000000} \textbf{0.023}}          & 3.502                                          & 0.058                                         \\ \hline
fr2\_desk                       & 1.060                                   & 0.030                                   & 0.463                                              & \textbf{0.008}                                    & 0.469                                          & \textbf{0.008}                                 & \textbf{0.458}                                 & \textbf{0.008}                                \\ \hline
fr2\_desk\_with\_person         & 1.639                                   & 0.056                                   & 0.472                                              & 0.012                                             & \textbf{0.462}                                 & \textbf{0.008}                                 & 0.511                                          & 0.009                                         \\ \hline
fr2\_dishes                     & 1.624                                   & 0.035                                   & 0.653                                              & \textbf{0.012}                                    & 0.915                                          & 0.016                                          & \textbf{0.629}                                 & \textbf{0.012}                                \\ \hline
fr2\_rpy                        & 0.357                                   & 0.004                                   & 0.321                                              & 0.003                                             & \textbf{0.318}                                 & \textbf{0.003}                                 & 0.325                                          & 0.004                                         \\ \hline
fr2\_xyz                        & 0.328                                   & 0.005                                   & 0.317                                              & 0.004                                             & \textbf{0.307}                                 & \textbf{0.003}                                 & 0.319                                          & 0.004                                         \\ \hline
fr3\_cabinet                    & 2.976                                   & 0.071                                   & 2.024                                              & 0.040                                             & 2.482                                          & 0.058                                          & \textbf{1.636}                                 & \textbf{0.036}                                \\ \hline
fr3\_large\_cabinet             & \textbf{2.369}                          & \textbf{0.100}                          & 4.717                                              & 0.214                                             & 4.036                                          & 0.190                                          & {\color[HTML]{000000} 3.278}                   & {\color[HTML]{000000} 0.167}                  \\ \hline
fr3\_long\_office\_household    & 0.906                                   & 0.019                                   & 0.529                                              & 0.011                                             & {\color[HTML]{000000} 0.695}                   & {\color[HTML]{000000} 0.014}                   & {\color[HTML]{000000} \textbf{0.503}}          & {\color[HTML]{000000} \textbf{0.010}}         \\ \hline
fr3\_nostr\_tex\_far            & 2.449                                   & 0.121                                   & 1.306                                              & 0.054                                             & {\color[HTML]{000000} 9.412}                   & {\color[HTML]{000000} 0.522}                   & \textbf{0.892}                                 & \textbf{0.035}                                \\ \hline
fr3\_nostr\_tex\_near\_withloop & 1.591                                   & 0.050                                   & 7.193                                              & 0.164                                             & {\color[HTML]{000000} \textbf{1.440}}          & {\color[HTML]{000000} \textbf{0.029}}          & {\color[HTML]{000000} 1.502}                   & {\color[HTML]{000000} 0.043}                  \\ \hline
fr3\_str\_notex\_far            & \textbf{0.453}                          & \textbf{0.013}                          & 1.935                                              & 0.132                                             & 3.104                                          & 0.144                                          & 0.588                                          & 0.027                                         \\ \hline
fr3\_str\_notex\_near           & \textbf{3.088}                          & \textbf{0.060}                          & 32.288                                             & 0.622                                             & {\color[HTML]{000000} 23.482}                  & {\color[HTML]{000000} 0.422}                   & 25.888                                         & 0.752                                         \\ \hline
fr3\_str\_tex\_far              & 0.618                                   & 0.018                                   & 0.472                                              & 0.013                                             & \textbf{0.459}                                 & \textbf{0.012}                                 & 0.477                                          & 0.013                                         \\ \hline
fr3\_str\_tex\_near             & 0.890                                   & 0.017                                   & 1.102                                              & 0.018                                             & {\color[HTML]{000000} 1.167}                   & {\color[HTML]{000000} 0.021}                   & \textbf{0.593}                                 & \textbf{0.010}                                \\ \hline
\end{tabular}
\end{table*}

\begin{table*}[p]
\centering
\caption{Absolute Trajectory RMSE(m) of TUM datasets}
\label{table: TUM ATE evaluation}
\begin{tabular}{|c|c|l|c|l|c|c|}
\hline
\multicolumn{1}{|l|}{}          & \multicolumn{2}{c|}{\begin{tabular}[c]{@{}c@{}}ORB-SLAM2\\ Features\end{tabular}} & \multicolumn{2}{c|}{\begin{tabular}[c]{@{}c@{}}Our Implementation (EDT)\\ Edge Alignment\end{tabular}} & \begin{tabular}[c]{@{}c@{}}Our Method (ANNF)\\ Edge Alignment\end{tabular} & \begin{tabular}[c]{@{}c@{}}Our Method (ONNF)\\ Edge Alignment\end{tabular} \\ \hline
Seq.                            & \multicolumn{2}{c|}{RMSE(t)}                                                      & \multicolumn{2}{c|}{RMSE(t)}                                                                           & RMSE(t)                                                                    & RMSE(t)                                                                   \\ \hline
fr1\_360                        & \multicolumn{2}{c|}{\textbf{0.139}}                                               & \multicolumn{2}{c|}{0.607}                                                                             & 0.315                                                                      & 0.242                                                                     \\ \hline
fr1\_desk                       & \multicolumn{2}{c|}{0.065}                                                        & \multicolumn{2}{c|}{0.168}                                                                             & {\color[HTML]{000000} 0.212}                                               & \textbf{0.044}                                                            \\ \hline
fr1\_desk2                      & \multicolumn{2}{c|}{\textbf{0.093}}                                               & \multicolumn{2}{c|}{0.581}                                                                             & 0.381                                                                      & 0.187                                                                     \\ \hline
fr1\_floor                      & \multicolumn{2}{c|}{0.061}                                                        & \multicolumn{2}{c|}{0.019}                                                                             & \textbf{0.017}                                                             & 0.021                                                                     \\ \hline
fr1\_plant                      & \multicolumn{2}{c|}{0.067}                                                        & \multicolumn{2}{c|}{\textbf{0.042}}                                                                    & 0.133                                                                      & 0.059                                                                     \\ \hline
fr1\_room                       & \multicolumn{2}{c|}{\textbf{0.143}}                                               & \multicolumn{2}{c|}{0.248}                                                                             & 0.621                                                                      & 0.242                                                                     \\ \hline
fr1\_rpy                        & \multicolumn{2}{c|}{0.066}                                                        & \multicolumn{2}{c|}{0.109}                                                                             & 0.205                                                                      & \textbf{0.047}                                                            \\ \hline
fr1\_teddy                      & \multicolumn{2}{c|}{\textbf{0.150}}                                               & \multicolumn{2}{c|}{0.290}                                                                             & 0.437                                                                      & 0.193                                                                     \\ \hline
fr1\_xyz                        & \multicolumn{2}{c|}{\textbf{0.009}}                                               & \multicolumn{2}{c|}{0.053}                                                                             & 0.137                                                                      & 0.043                                                                     \\ \hline
fr2\_360\_hemisphere            & \multicolumn{2}{c|}{0.213}                                                        & \multicolumn{2}{c|}{0.504}                                                                             & 0.432                                                                      & \textbf{0.079}                                                            \\ \hline
fr2\_360\_kidnap                & \multicolumn{2}{c|}{0.144}                                                        & \multicolumn{2}{c|}{0.249}                                                                             & 0.128                                                                      & \textbf{0.122}                                                            \\ \hline
fr2\_coke                       & \multicolumn{2}{c|}{1.521}                                                        & \multicolumn{2}{c|}{0.076}                                                                             & {\color[HTML]{000000} \textbf{0.029}}                                      & 0.070                                                                     \\ \hline
fr2\_desk                       & \multicolumn{2}{c|}{0.274}                                                        & \multicolumn{2}{c|}{0.040}                                                                             & 0.039                                                                      & \textbf{0.037}                                                            \\ \hline
fr2\_desk\_with\_person         & \multicolumn{2}{c|}{0.135}                                                        & \multicolumn{2}{c|}{0.072}                                                                             & \textbf{0.047}                                                             & 0.069                                                                     \\ \hline
fr2\_dishes                     & \multicolumn{2}{c|}{0.104}                                                        & \multicolumn{2}{c|}{0.035}                                                                             & 0.034                                                                      & \textbf{0.033}                                                            \\ \hline
fr2\_rpy                        & \multicolumn{2}{c|}{\textbf{0.004}}                                               & \multicolumn{2}{c|}{0.009}                                                                             & 0.007                                                                      & 0.007                                                                     \\ \hline
fr2\_xyz                        & \multicolumn{2}{c|}{\textbf{0.008}}                                               & \multicolumn{2}{c|}{0.009}                                                                             & 0.010                                                                      & \textbf{0.008}                                                            \\ \hline
fr3\_cabinet                    & \multicolumn{2}{c|}{0.312}                                                        & \multicolumn{2}{c|}{\textbf{0.057}}                                                                    & 0.103                                                                      & \textbf{0.057}                                                            \\ \hline
fr3\_large\_cabinet             & \multicolumn{2}{c|}{\textbf{0.154}}                                               & \multicolumn{2}{c|}{0.351}                                                                             & 0.349                                                                      & 0.317                                                                     \\ \hline
fr3\_long\_office\_household    & \multicolumn{2}{c|}{0.276}                                                        & \multicolumn{2}{c|}{0.087}                                                                             & 0.090                                                                      & \textbf{0.085}                                                            \\ \hline
fr3\_nostr\_tex\_far            & \multicolumn{2}{c|}{0.147}                                                        & \multicolumn{2}{c|}{0.055}                                                                             & 0.191                                                                      & \textbf{0.026}                                                            \\ \hline
fr3\_nostr\_tex\_near\_withloop & \multicolumn{2}{c|}{0.111}                                                        & \multicolumn{2}{c|}{0.406}                                                                             & 0.101                                                                      & \textbf{0.090}                                                            \\ \hline
fr3\_str\_notex\_far            & \multicolumn{2}{c|}{\textbf{0.008}}                                               & \multicolumn{2}{c|}{0.157}                                                                             & 0.026                                                                      & 0.031                                                                     \\ \hline
fr3\_str\_notex\_near           & \multicolumn{2}{c|}{\textbf{0.091}}                                               & \multicolumn{2}{c|}{0.910}                                                                             & 0.813                                                                      & 1.363                                                                     \\ \hline
fr3\_str\_tex\_far              & \multicolumn{2}{c|}{0.030}                                                        & \multicolumn{2}{c|}{0.013}                                                                             & \textbf{0.012}                                                             & 0.013                                                                     \\ \hline
fr3\_str\_tex\_near             & \multicolumn{2}{c|}{0.045}                                                        & \multicolumn{2}{c|}{0.026}                                                                             & 0.047                                                                      & \textbf{0.025}                                                            \\ \hline
\end{tabular}
\end{table*}

\begin{figure*}[p]
  \centering
  \subfigure[\textit{fr2\_xyz}.]{
  \includegraphics[width=0.40\textwidth]{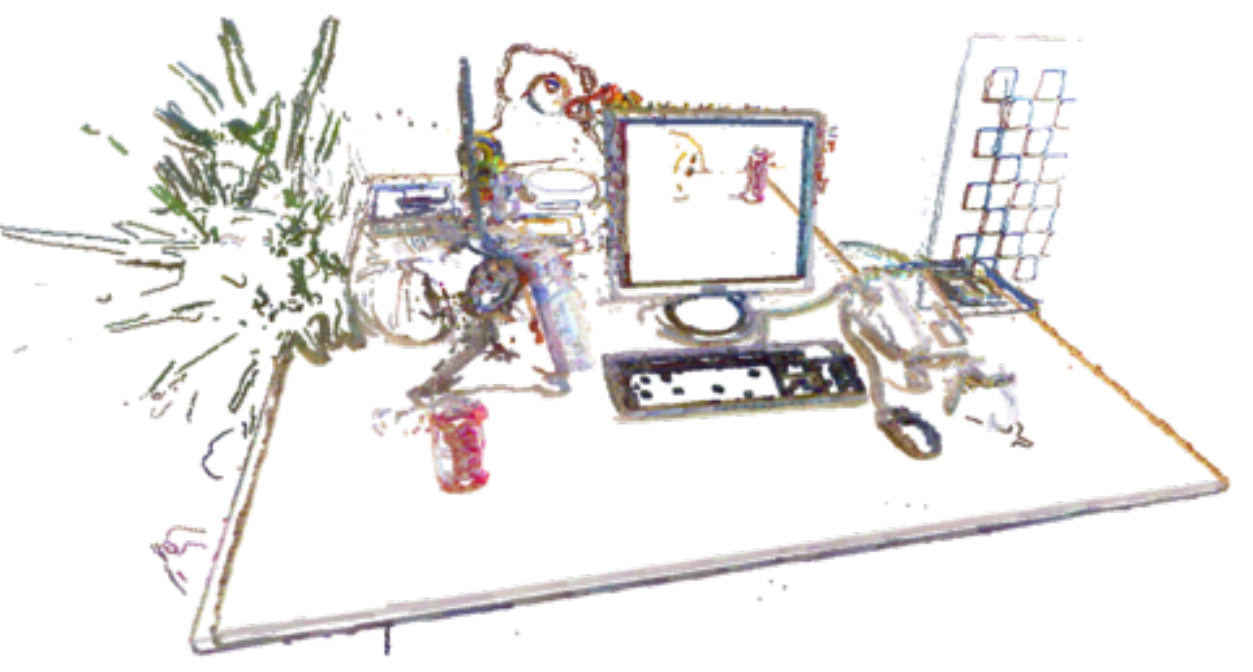}}
  \subfigure[\textit{fr3\_nostructure\_texture\_near}.]{
  \includegraphics[width=0.28\textwidth]{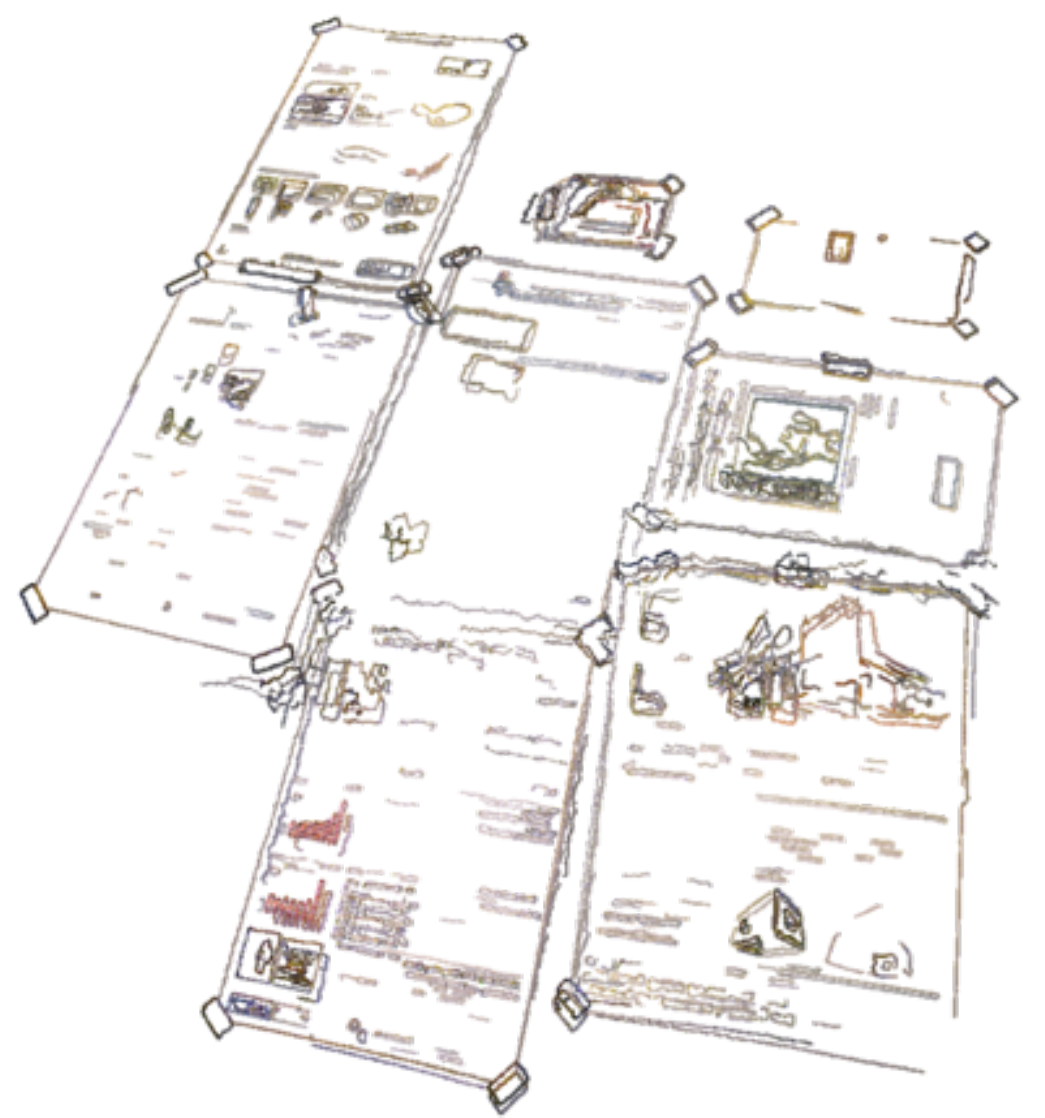}}
  \caption{Semi-dense reconstruction of sequence \textit{freiburg2\_xyz} and \textit{freiburg3\_nostructure\_texture\_near} from the TUM RGB-D benchmark datasets.}
  \label{fig: tum reconstruction}
\end{figure*}

As shown in Tables~\ref{table: TUM RPE evaluation} and ~\ref{table: TUM ATE evaluation}, the ANNF based paradigm achieves better accuracy than EDF (which we attribute to the use of the signed point-to-tangent distance), and ONNF based tracking significantly outperforms other methods due to bias-free estimation. Since edge alignment methods rely on accurate Canny edge detections, it is not surprising to see~\cite{murORB2} performs better on several sequences in \textit{freiburg 1}, in which significant image blur due to aggressive rotations occurs. This problem would be less apparent if using a more advanced device,~\eg~Kinect V2, which is equipped with a global shutter RGB camera. Large RMSEs of edge alignment based methods are also witnessed in other sequences such as \textit{fr3\_structure\_texture\_near}, which is caused by an ambiguous structure. Only one edge is detected in the conjunction of two planes with homogeneous color, which notably leads to a tracking failure, as at least one degree of freedom of the motion simply becomes unobservable\footnote{Note furthermore that---in order to achieve a fair comparison---all methods are evaluated up until the same frame if one of the methods loses tracking. Since all the edge alignment based methods are equally affectable by failure situations, this does not give preference to any of the methods.}. In general, however, ANNF and ONNF based trackers work outstandingly well, since the median errors remain reasonably small. To conclude, semi-dense reconstruction results for the sequences \textit{fr2\_xyz} and \textit{fr3\_nostructure\_texture} are given in Fig.~\ref{fig: tum reconstruction}. Since no global optimization is performed, the crispness of these reconstructions again underlines the quality of the edge alignment.

\subsection{ICL-NUIM Dataset}
\label{subsec:ICL dataset}
\begin{figure*}[p]
  \centering
  \subfigure[Aerial view of the living room.]{
  \includegraphics[width=0.3\textwidth]{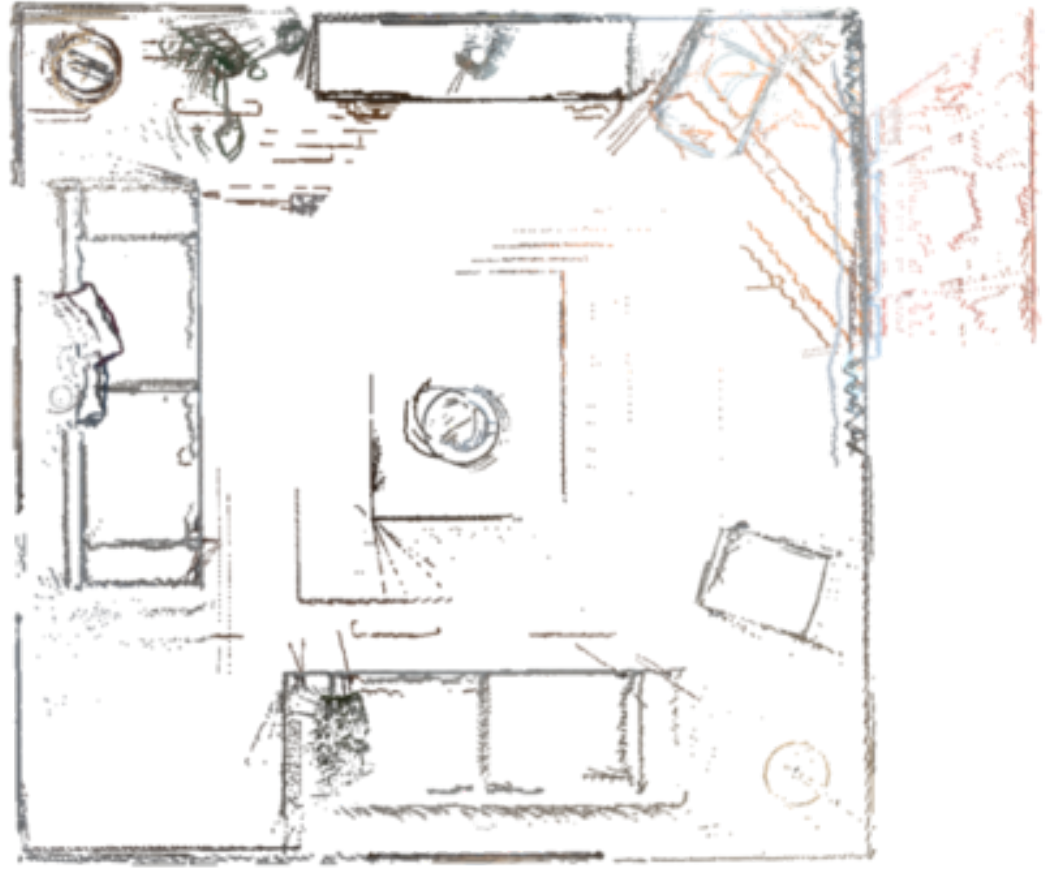}}
  \subfigure[Horizontal view of the living room.]{
  \includegraphics[width=0.3\textwidth]{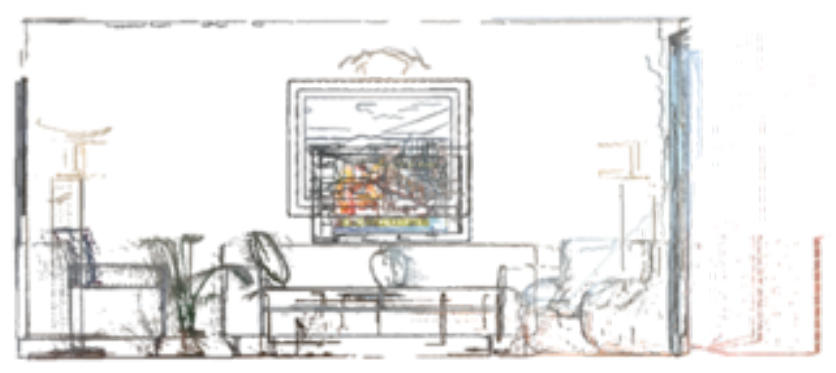}}
  \subfigure[A manually selected perspective.]{
  \includegraphics[width=0.3\textwidth]{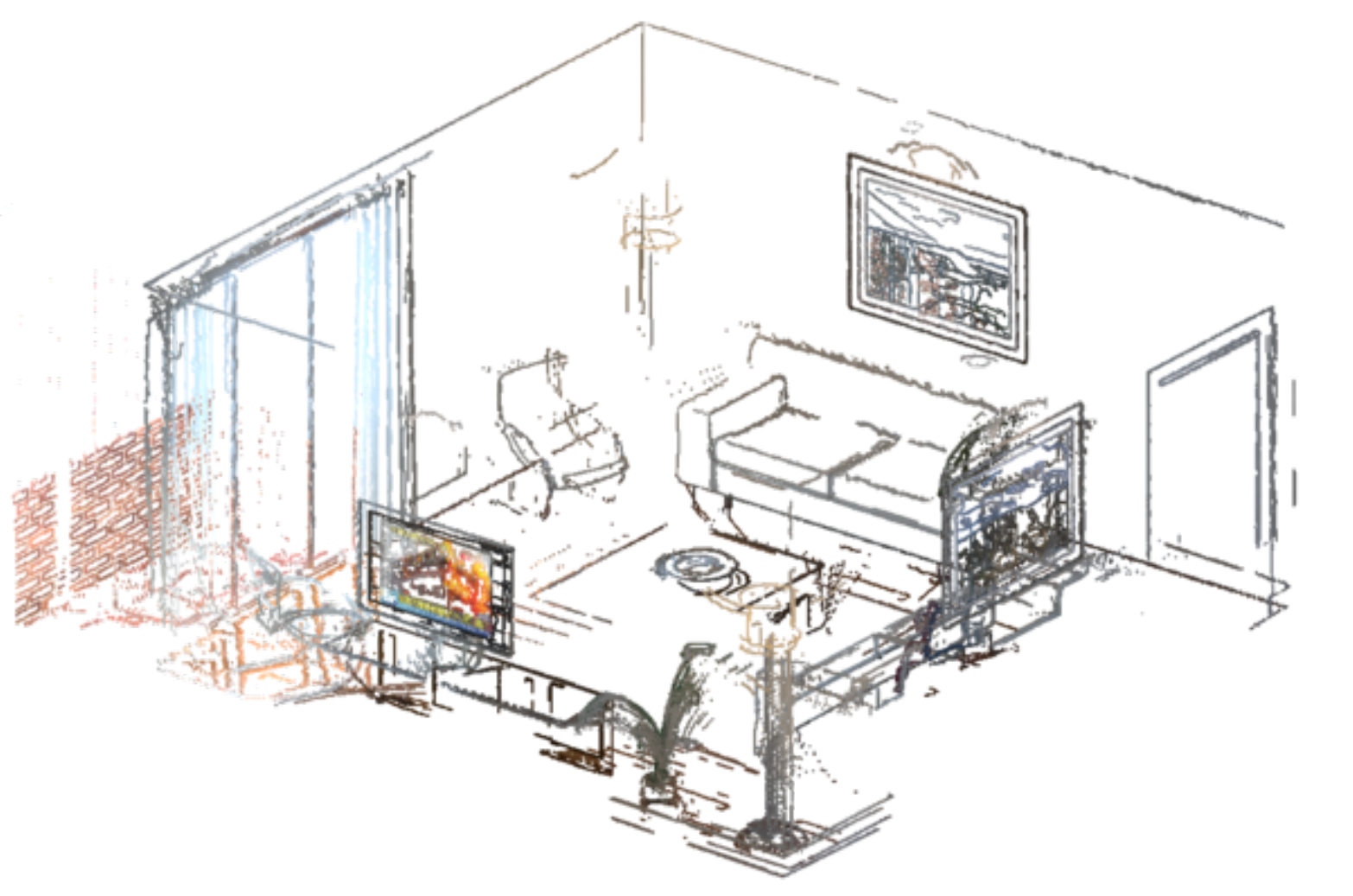}}
 
  \caption{Semi-dense reconstruction of sequence \textit{living room kt2} from the ICL\_NUIM dataset. Views at different perspectives are provided.}
  \label{fig: ICL NUIM reconstruction}
\end{figure*}

A high-quality indoor dataset for evaluating RGB-D VO/SLAM systems is provided by Handa in 2014~\cite{handa:etal:ICRA2014}. Although it is synthetic, the structure and texture are realistically rendered using a professional 3D content creation software. Illumination and reflection properties are properly taken into account. We evaluate our algorithm using the \textit{living room} collection which contains four sequences composed of different trajectories within the same room. The scene has several challenging elements for VO/SLAM systems, including reflective surfaces, locally texture-poor regions, and multiple illumination sources. The evaluation results are given in Table~\ref{table: RPE of ICL} and~\ref{table: ATE of ICL}. We see that the ONNF based tracker again gives the best performance in the comparison. Since image blur effects do not exist in the synthetic dataset, the advantages of the ONNF based tracking scheme are even more clearly demonstrated. The performance of ORB-SLAM2 is affected by locally textureless scenarios at some points, where only blobs and curves (boundaries of objects) appear rather than corner-like features. To conclude, we again provide a semi-dense reconstruction of the \textit{living room kt2} using ONNF based tracking in Fig.~\ref{fig: ICL NUIM reconstruction}.

\begin{table*}[p]
\centering
\caption{Relative pose RMSE $\mathbf{R}$:deg/s, $\mathbf{t}$:m/s of ICL\_NUIM }
\label{table: RPE of ICL}
\begin{tabular}{|c|c|c|c|c|c|c|c|c|}
\hline
              & \multicolumn{2}{c|}{\begin{tabular}[c]{@{}c@{}}ORB-SLAM2\\ Features\end{tabular}} & \multicolumn{2}{c|}{\begin{tabular}[c]{@{}c@{}}Our Implementation (EDF)\\ Edge Alignment\end{tabular}} & \multicolumn{2}{c|}{\begin{tabular}[c]{@{}c@{}}Our Method (ANNF)\\ Edge Alignment\end{tabular}} & \multicolumn{2}{c|}{\begin{tabular}[c]{@{}c@{}}Our Method (ONNF)\\ Edge Alignment\end{tabular}} \\ \hline
Seq.          & RMSE(R)                                 & RMSE(t)                                 & RMSE(R)                                            & RMSE(t)                                           & RMSE(R)                                     & RMSE(t)                                           & RMSE(R)                                 & \multicolumn{1}{l|}{RMSE(t)}                         \\ \hline
living room 0 & 1.186                                   & 0.030                                   & 2.717                                              & 0.082                                             & 1.766                                       & 0.047                                             & \textbf{0.674}                          & \textbf{0.014}                                       \\ \hline
living room 1 & 0.464                                   & 0.022                                   & 0.590                                              & 0.030                                             & 1.297                                       & 0.059                                             & \textbf{0.208}                          & \textbf{0.009}                                       \\ \hline
living room 2 & 2.997                                   & 0.103                                   & 0.544                                              & 0.029                                             & 0.307                                       & 0.013                                             & \textbf{0.269}                          & \textbf{0.011}                                       \\ \hline
living room 3 & 0.367                                   & 0.012                                   & 0.214                                              & 0.011                                             & 0.157                                       & \textbf{0.007}                                    & \textbf{0.152}                          & \textbf{0.007}                                       \\ \hline
\end{tabular}
\end{table*}

\begin{table*}[p]
\centering
\renewcommand{\arraystretch}{1.2}
\caption{Absolute Trajectory RMSE (m) of ICL\_NUIM}
\label{table: ATE of ICL}
\begin{tabular}{|c|c|c|c|c|}
\hline
              & \begin{tabular}[c]{@{}c@{}}ORB-SLAM2\\ Features\end{tabular} & \begin{tabular}[c]{@{}c@{}}Our Implementation (EDF)\\ Edge Alignment\end{tabular} & \begin{tabular}[c]{@{}c@{}}Our Method (ANNF)\\ Edge Alignment\end{tabular} & \begin{tabular}[c]{@{}c@{}}Our Method (ONNF)\\ Edge Alignment\end{tabular} \\ \hline
Seq.          & RMSE(t)                                                      & RMSE(t)                                                                           & RMSE(t)                                                                    & RMSE(t)                                                                   \\ \hline
living room 0 & 0.043                                                        & 0.113                                                                             & 0.074                                                                      & \textbf{0.035}                                                            \\ \hline
living room 1 & 0.082                                                        & 0.080                                                                             & 0.119                                                                      & \textbf{0.023}                                                            \\ \hline
living room 2 & 0.108                                                        & 0.089                                                                             & 0.038                                                                      & \textbf{0.031}                                                            \\ \hline
living room 3 & 0.015                                                        & 0.016                                                                             & \textbf{0.008}                                                             & \textbf{0.008}                                                            \\ \hline
\end{tabular}
\end{table*}

\begin{figure*}[p]
  \centering  
  \subfigure[Floorplan of level 3 of the ANU research school of engineering.]{
  \includegraphics[width=\columnwidth]{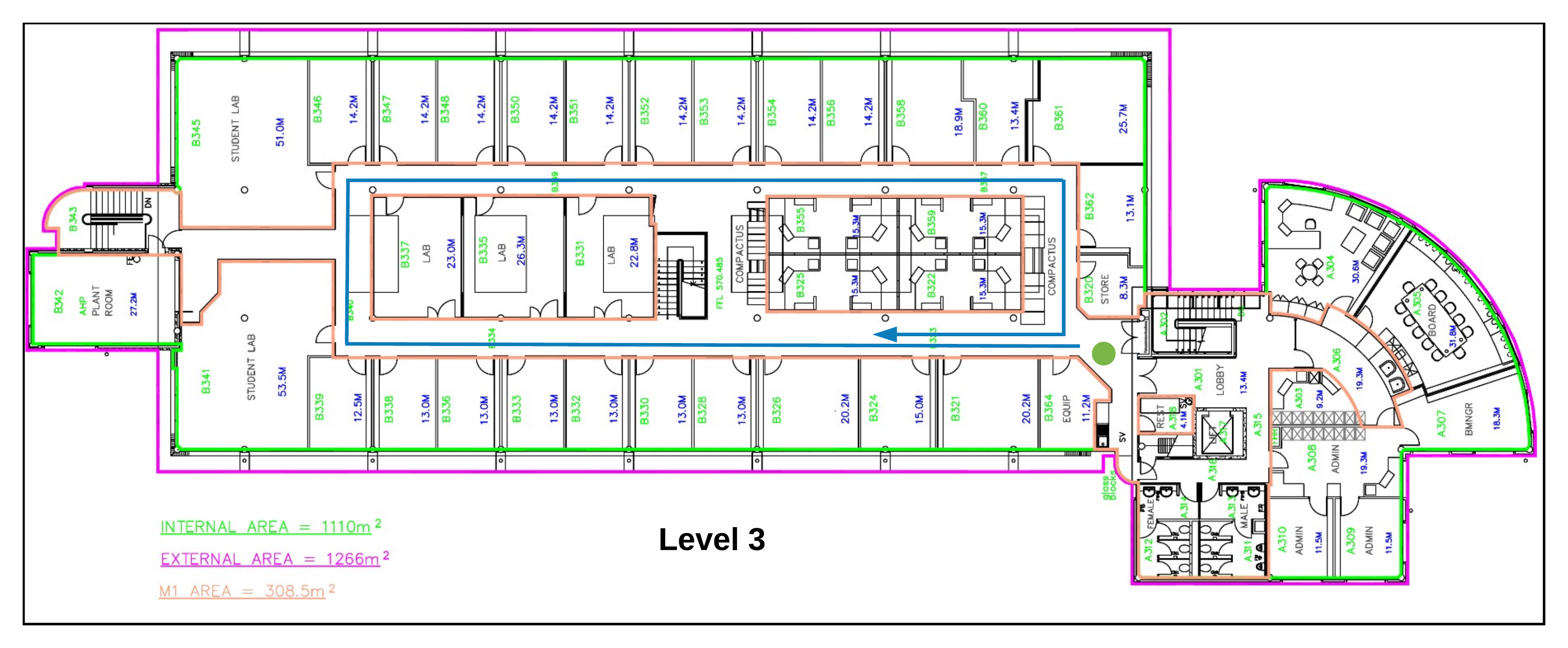}}
  \subfigure[Typical snapshots of the environment.]{
  \includegraphics[width=\columnwidth]{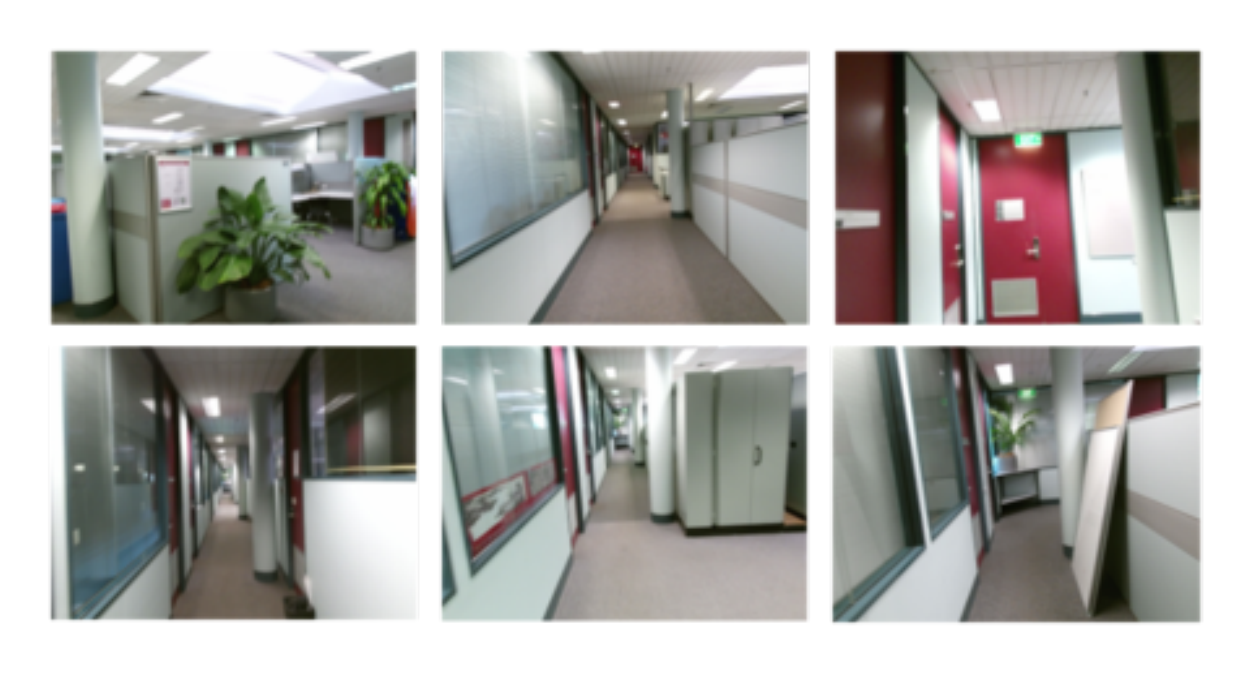}}
  \caption{The schematic trajectory of the sensor when collecting the sequence is illustrated in (a). The sequence starts from the position highlighted with a green dot. Structures such as window glass, plants and dark corridor caused by inconsistent illumination make the sequence challenging are shown in (b).}
  \label{fig:Brian Anderson building}
\end{figure*}
\begin{figure*}[p]
  \centering
  \subfigure[Reconstruction result by ORB-SLAM2 (RGB-D version).]{
  \includegraphics[width=0.43\textwidth]{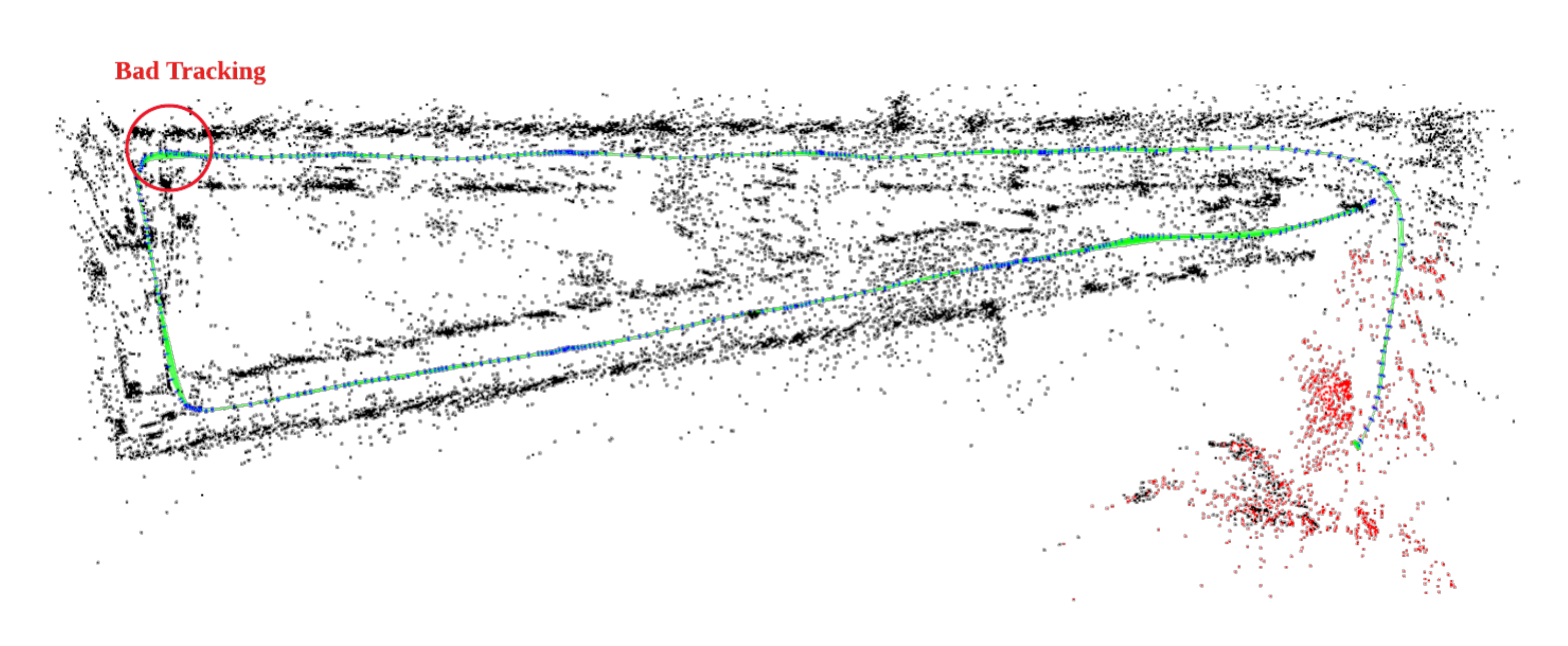}}
  \subfigure[Reconstruction result by smoothed EDF based tracker.]{
  \includegraphics[width=0.43\textwidth]{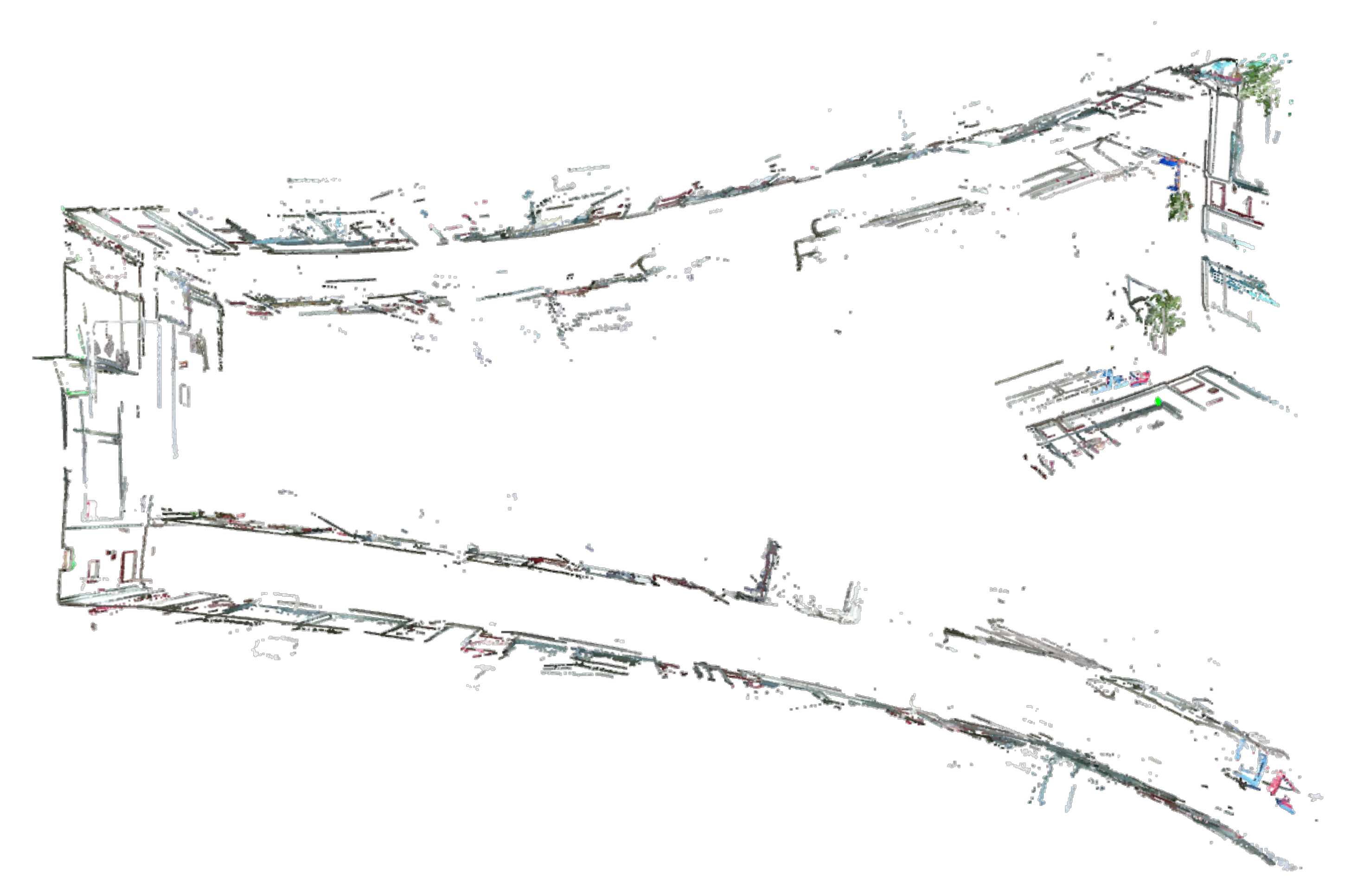}}
  \subfigure[Reconstruction result by ANNF based tracker.]{
  \includegraphics[width=0.43\textwidth]{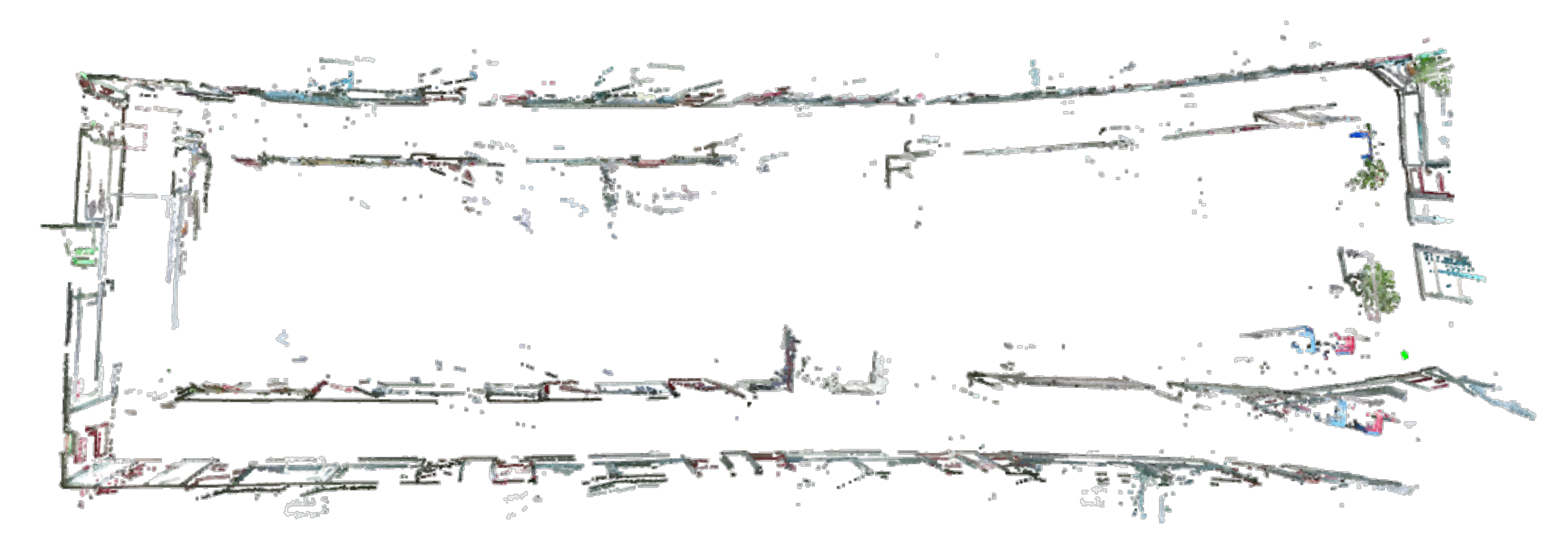}}
  \subfigure[Reconstruction result by ONNF based tracker.]{
  \includegraphics[width=0.43\textwidth]{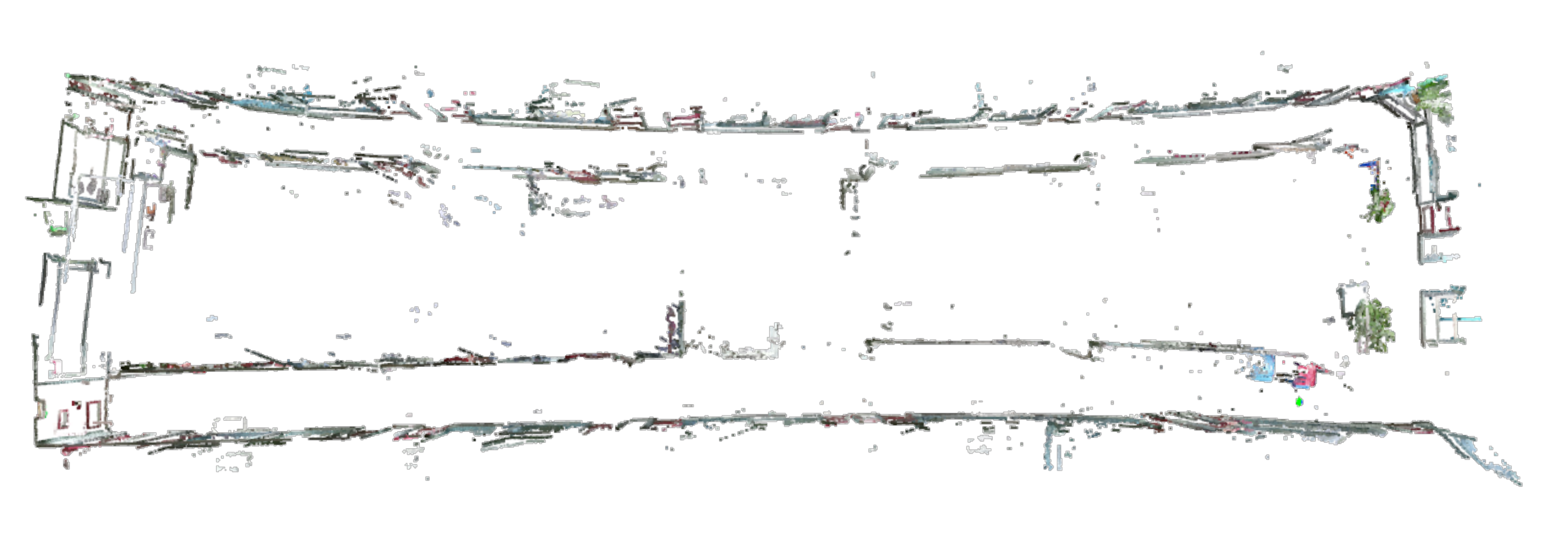}}
  \caption{Qualitative evaluation on the ANU-RSISE RGB-D sequence. The reconstruction results of four methods are provided respectively.}
  \label{fig:RSISE Reconstruction comparison}
\end{figure*}
\begin{figure*}[p]
  \centering
  \includegraphics[width=0.7\textwidth]{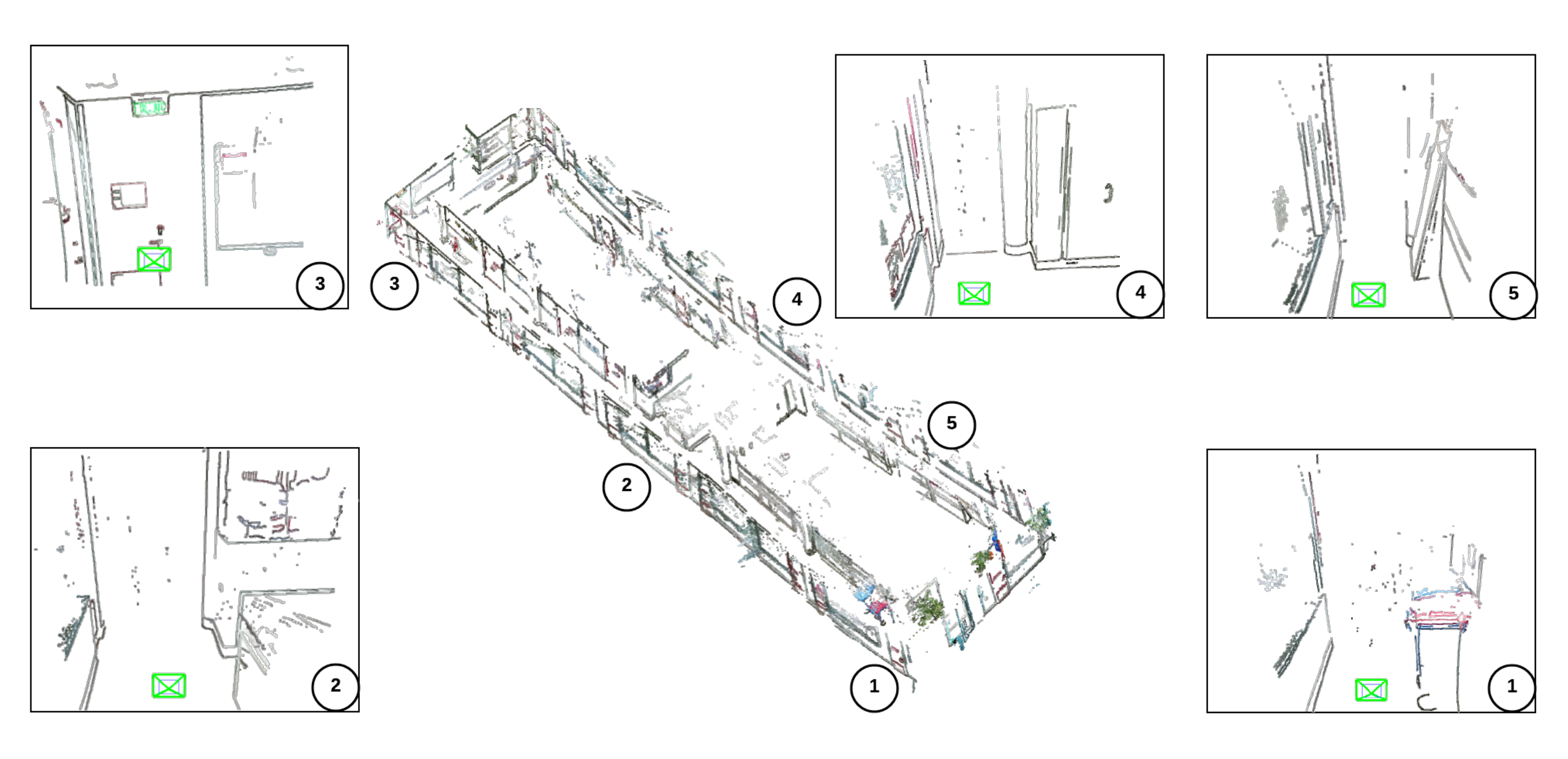}
  \caption{Close-up perspectives during the exploration of level 3 of the ANU research school of engineering.}
  \label{fig:RSISE Reconstruction 2}
\end{figure*}
\subsection{ANU-RSISE Sequence}

We captured and analyzed our own large-scale indoor RGB-D sequence, a scan of a complete level of the Research School of Engineering at the Australian National University (ANU). It is more challenging than most of the TUM datasets for at least two reasons. First, the scene is an open-space office area of approximately 300 $m^2$, with highly self-similar locations. A footprint of the building is shown in Fig.~\ref{fig:Brian Anderson building}. The illumination is not as consistent as in small-scale environments, such as a desk or a small office room. Second, the sequence contains a combination of challenging structures such as reflecting surfaces (window glass) and cluttered objects (plants). We use a Microsoft Kinect v2 for data collection, and the RGB and depth images are prealigned and resized to VGA resolution, similar to what has been done in the TUM benchmark sequences.

All algorithms are evaluated qualitatively by visualizing the reconstruction results in Fig.~\ref{fig:RSISE Reconstruction comparison}. The global BA module of~\cite{murORB2} is again disabled to underline pure tracking performance. Although~\cite{murORB2} performs very well along straight parts, severe problems are witnessed in the corners. The bad tracking is due to the reflectance imaging on the window glass, which generates false features. All edge alignment based tracker still perform well in the corner taking advantage of good signal-to-noise ratio and the proposed robust weighting strategies. The advantages of the ANNF and ONNF over the EDF are clearly seen over the straight parts. By looking at the two recycle bins (blue and red) near the starting point, ONNF performs the best in terms of start-to-end error. Note that the straight corridors look slightly bended because of some see-through effects on the side with transparent window glass, which provide inaccurate depth measurements. A more detailed map and some close-up shots occurring during the exploration using ONNF based tracking are given in Fig.~\ref{fig:RSISE Reconstruction 2}.

\subsection{Efficiency Analysis}
\label{subsec: efficiency analysis}
\begin{figure}[H]
  \centering
  \includegraphics[width=0.4\textwidth]{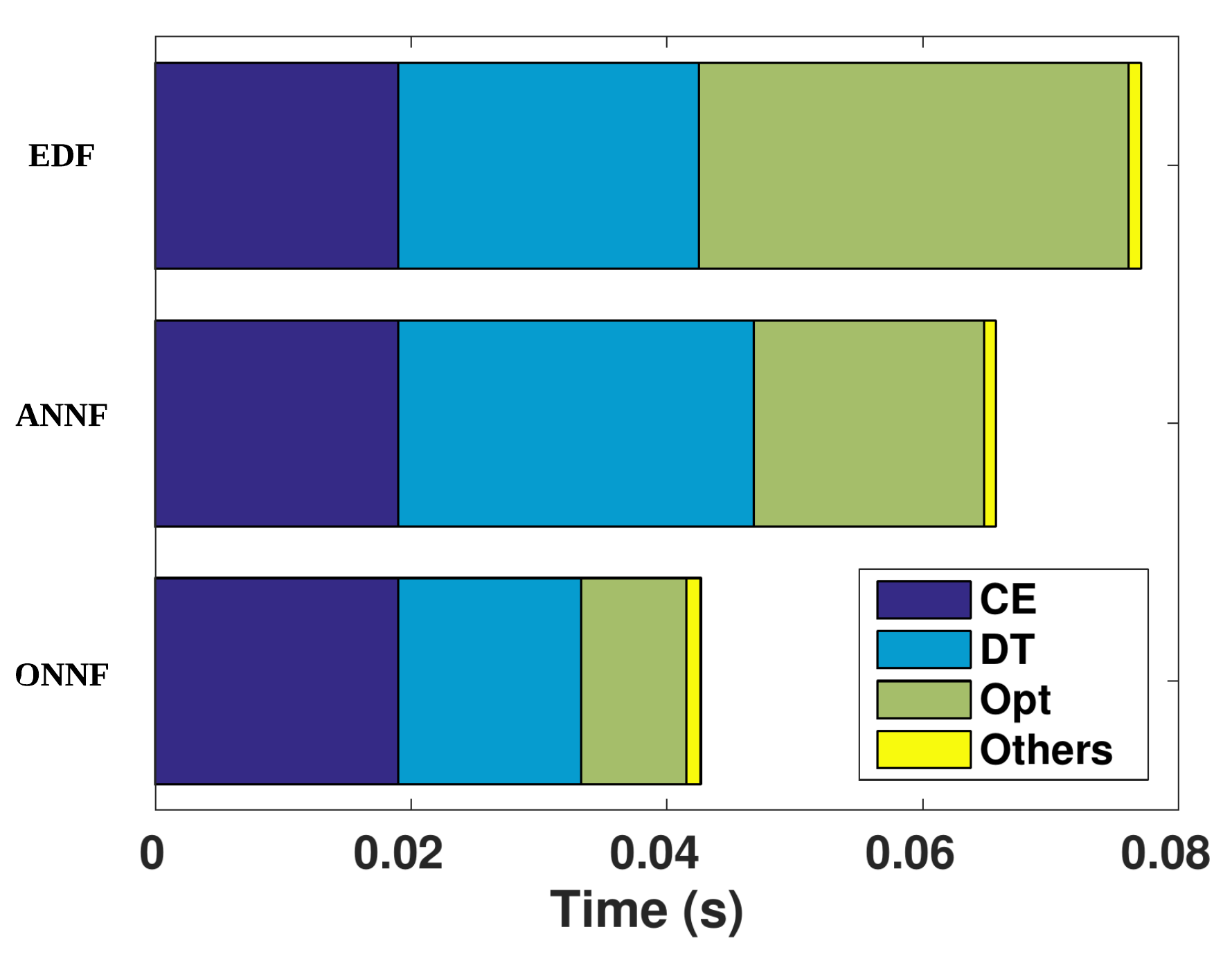}
  \caption{Efficiency analysis on EDF, ANNF and ONNF based tracker.}
  \label{fig:efficiency bar}
\end{figure}

Real-time performance is typically required for any VO system in a practical application. To see the improvement in terms of efficiency, we compare the computation time of each method on a desktop with a Core i7-4770 CPU. As seen in Fig.~\ref{fig:efficiency bar}, the computation in the tracking thread consists of four parts: Canny edge detection (CE), distance transformation (DT), optimization (Opt), and others. As claimed before, the DT computation of the ANNF\footnote{The adaptive sampling function is switched off for ANNF in the efficiency analysis.} is almost as fast as the EDF, while the ONNF is the most efficient due to the adaptive sampling and the parallel computation. Another significant difference occurs in the optimization. The EDF based method takes more time than the ANNF because of the requirement for bilinear interpolation during the evaluation of the objective function. ONNF based tracking is basically as fast as ANNF based tracking. The difference in the optimization time for nearest neighbour field based approaches is due to another modification. 
We include a stochastic optimization strategy in
\noindent the implementation of ONNF based tracking, which starts with a small number of 3D points and gradually increases the amount until reaching the minimum, where optimization over all points is performed. Note that the result in Fig.~\ref{fig:efficiency bar} is normalized over the number of points (at most 6500) and it includes the computation on the whole image pyramid (from level 0 to level 2). Additionally, the reference frame preparation thread runs at 10 Hz in parallel.

Even using three pyramid levels, our method achieves 25 Hz approximately and thus real-time processing on a standard CPU. The main bottleneck in the computation is the image processing. Considering that this could be offloaded into embedded hardware, we believe that our method represents an interesting choice for computationally constrained devices.
\section{Conclusion}
\label{sec:conclusion}

The present paper introduces approximate nearest neighbour fields as a valid, at least equally accurate alternative to euclidean distance fields in 3D-2D curve alignment with clear benefits in computational efficiency. We furthermore prove that the bias plaguing distance field based registration in the case of partially observed models is effectively encountered through an orientation of the nearest neighbour fields, thus reestablishing the model-to-data registration paradigm as the most efficient choice for geometric 3D-2D curve alignment. We furthermore prove that efficient sub-sampling strategies are readily accessible to nearest neighbour field extraction.

The geometric approach to semi-dense feature-based alignment has the clear advantages of resilience to illumination changes and the ability to be included in a curve-based bundle adjustment that relies on a global, spline-based representation of the structure. With a focus on the efficient formulation of residual errors in curve alignment, we believe that the present investigation represents an important addition to this line of research. Our future objectives consist of including oriented nearest neighbour fields into large-scale edge-based SLAM and a spline-based global optimization framework for regular cameras.

\newpage
\section{Appendix}
\subsection{Derivation on Jacobian Matrix of ANNF based Tracking}
\label{subsec: appendix derivation of J}

The linearization of the residual function at $\mathbold{\theta}_k$ is
\begin{equation}
r_{lin,i}(\mathbold{\theta}_{k+1}) = r_{i}(\mathbold{\theta}_{k}) + \mathbf{J}_{i}(\mathbold{\theta}_{k})\Delta\mathbold{\theta}.
\end{equation}

The Jacobian matrix could be obtained using chains rule as

\begin{equation}
\mathbf{J}_{i}(\mathbold{\theta}_{k})=g(\mathbf{x}_i)^{T}\mathbf{J}_{\pi}\mathbf{J}_{T}\mathbf{J}_{G}.
\end{equation}

Each sub Jacobian matrix are derived as following.

\begin{equation}
\mathbf{J}_{\pi} = \frac{\partial \pi}{\partial T}\vert_{\mathbf{p} = T(G(\mathbold{\theta}_{k}),\mathbf{x}_{i})}= \left[\begin{matrix}
f_x\frac{1}{z^{\prime}} & 0 & -f_x\frac{x^{\prime}}{{z^{\prime}}^{2}} \\
0 & f_y\frac{1}{z^{\prime}} & -f_y\frac{y^{\prime}}{{z^{\prime}}^{2}}
\end{matrix}\right],
\end{equation}
where $\mathbf{p}_i^{\prime} = (x^{\prime},y^{\prime},z^{\prime})$ is the 3D point transformed by motion $G(\mathbold{\theta}_{k})$.

\begin{align}
\label{eq: J_T fam}
\mathbf{J}_{T} &= \frac{\partial T}{\partial G}\vert_{G=G(\mathbold{\theta}_{k}), \mathbf{p} = \mathbf{p}_i} \\ \nonumber
&= \left[ \begin{array}{@{}*{12}{c}@{}}
     x & 0 & 0 & y & 0 & 0 & z & 0 & 0 & 1 & 0 & 0\\
     0 & x & 0 & 0 & y & 0 & 0 & z & 0 & 0 & 1 & 0\\
     0 & 0 & x & 0 & 0 & y & 0 & 0 & z & 0 & 0 & 1
\end{array}
\right].
\end{align}

$\mathbf{J}_G$ can be obtained by computing the derivatives of the pose $G$ with respect to the motion parameter $\mathbold{\theta} = \left[ t_1, t_2, t_3, c_1, c_2, c_3 \right]^{T}$, shown as below

\begin{align}
\label{eq: J_G fam cayley}
\mathbf{J}_{G} &=
\left[ \begin{matrix}%
     \frac{\partial r_{11}}{\partial t_{1}} & \frac{\partial r_{11}}{\partial t_{2}} & \frac{\partial r_{11}}{\partial t_{3}} & \frac{\partial r_{11}}{\partial c_{1}} & \frac{\partial r_{11}}{\partial c_{2}} & \frac{\partial r_{11}}{\partial c_{3}} \\
     \frac{\partial r_{21}}{\partial t_{1}} & \frac{\partial r_{21}}{\partial t_{2}} & \frac{\partial r_{21}}{\partial t_{3}} & \frac{\partial r_{21}}{\partial c_{1}} & \frac{\partial r_{21}}{\partial c_{2}} & \frac{\partial r_{21}}{\partial c_{3}} \\
     \frac{\partial r_{31}}{\partial t_{1}} & \frac{\partial r_{31}}{\partial t_{2}} & \frac{\partial r_{31}}{\partial t_{3}} & \frac{\partial r_{31}}{\partial c_{1}} & \frac{\partial r_{31}}{\partial c_{2}} & \frac{\partial r_{31}}{\partial c_{3}} \\
     \frac{\partial r_{12}}{\partial t_{1}} & \frac{\partial r_{12}}{\partial t_{2}} & \frac{\partial r_{12}}{\partial t_{3}} & \frac{\partial r_{12}}{\partial c_{1}} & \frac{\partial r_{12}}{\partial c_{2}} & \frac{\partial r_{12}}{\partial c_{3}} \\
     \frac{\partial r_{22}}{\partial t_{1}} & \frac{\partial r_{22}}{\partial t_{2}} & \frac{\partial r_{22}}{\partial t_{3}} & \frac{\partial r_{22}}{\partial c_{1}} & \frac{\partial r_{22}}{\partial c_{2}} & \frac{\partial r_{22}}{\partial c_{3}} \\
     \frac{\partial r_{32}}{\partial t_{1}} & \frac{\partial r_{32}}{\partial t_{2}} & \frac{\partial r_{32}}{\partial t_{3}} & \frac{\partial r_{32}}{\partial c_{1}} & \frac{\partial r_{32}}{\partial c_{2}} & \frac{\partial r_{32}}{\partial c_{3}} \\
     \frac{\partial r_{13}}{\partial t_{1}} & \frac{\partial r_{13}}{\partial t_{2}} & \frac{\partial r_{13}}{\partial t_{3}} & \frac{\partial r_{13}}{\partial c_{1}} & \frac{\partial r_{13}}{\partial c_{2}} & \frac{\partial r_{13}}{\partial c_{3}} \\
     \frac{\partial r_{23}}{\partial t_{1}} & \frac{\partial r_{23}}{\partial t_{2}} & \frac{\partial r_{23}}{\partial t_{3}} & \frac{\partial r_{23}}{\partial c_{1}} & \frac{\partial r_{23}}{\partial c_{2}} & \frac{\partial r_{23}}{\partial c_{3}} \\
     \frac{\partial r_{33}}{\partial t_{1}} & \frac{\partial r_{33}}{\partial t_{2}} & \frac{\partial r_{33}}{\partial t_{3}} & \frac{\partial r_{33}}{\partial c_{1}} & \frac{\partial r_{33}}{\partial c_{2}} & \frac{\partial r_{33}}{\partial c_{3}} \\
     \frac{\partial t_1}{\partial t_{1}} & \frac{\partial t_1}{\partial t_{2}} & \frac{\partial t_1}{\partial t_{3}} & \frac{\partial t_1}{\partial c_{1}} & \frac{\partial t_1}{\partial c_{2}} & \frac{\partial t_1}{\partial c_{3}} \\
     \frac{\partial t_2}{\partial t_{1}} & \frac{\partial t_2}{\partial t_{2}} & \frac{\partial t_2}{\partial t_{3}} & \frac{\partial t_2}{\partial c_{1}} & \frac{\partial t_2}{\partial c_{2}} & \frac{\partial t_2}{\partial c_{3}} \\
     \frac{\partial t_3}{\partial t_{1}} & \frac{\partial t_3}{\partial t_{2}} & \frac{\partial t_3}{\partial t_{3}} & \frac{\partial t_3}{\partial c_{1}} & \frac{\partial t_3}{\partial c_{2}} & \frac{\partial t_3}{\partial c_{3}}
\end{matrix}
\right]_{12 \times 6} \\
&= \left[
\begin{matrix}
\mathbf{O}_3 & \mathbf{A}_{1} \\
\mathbf{O}_3 & \mathbf{A}_{2} \\
\mathbf{O}_3 & \mathbf{A}_{3} \\
\mathbf{I}_3 & \mathbf{O}_{3 \times 3}
\end{matrix}
\right] \nonumber.
\end{align}

let's denote $K = 1 + c_1^{2} + c_2^{2} + c_3^{3}$, then the entries of the matrices $\mathbf{A}_{1}$ are,
\begin{itemize}
\item $\frac{\partial r_{11}}{\partial c_{1}} = \frac{2c_1}{K} - \frac{2c_{1}(1 + c^{2}_1 - c^{2}_2 - c^{2}_3)}{K^{2}}$,
\item $\frac{\partial r_{11}}{\partial c_{2}} = -\frac{2c_2}{K} - \frac{2c_{2}(1 + c^{2}_1 - c^{2}_2 - c^{2}_3)}{K^{2}}$,
\item $\frac{\partial r_{11}}{\partial c_{3}} = -\frac{2c_3}{K} - \frac{2c_{3}(1 + c^{2}_1 - c^{2}_2 - c^{2}_3)}{K^{2}}$,
\item $\frac{\partial r_{21}}{\partial c_{1}} = \frac{2c_2}{K} - \frac{4c_{1}(c_1c_2 + c_3)}{K^{2}}$,
\item $\frac{\partial r_{21}}{\partial c_{2}} = \frac{2c_1}{K} - \frac{4c_{2}(c_1c_2 + c_3)}{K^{2}}$,
\item $\frac{\partial r_{21}}{\partial c_{3}} = \frac{2}{K} - \frac{4c_{3}(c_1c_2 + c_3)}{K^{2}}$,
\item $\frac{\partial r_{31}}{\partial c_{1}} = \frac{2c_3}{K} - \frac{4c_{1}(c_1c_3 - c_2)}{K^{2}}$,
\item $\frac{\partial r_{31}}{\partial c_{2}} = -\frac{2}{K} + \frac{4c_{2}(c_1c_3 - c_2)}{K^{2}}$,
\item $\frac{\partial r_{31}}{\partial c_{3}} = \frac{2c_1}{K} - \frac{4c_{3}(c_1c_3 - c_2)}{K^{2}}$,
\end{itemize}

the entries of the matrices $\mathbf{A}_{2}$ are respectively,
\begin{itemize}
\item $\frac{\partial r_{12}}{\partial c_{1}} = \frac{2c_2}{K} - \frac{4c_{1}(c_1c_2 - c_3)}{K^{2}}$,
\item $\frac{\partial r_{12}}{\partial c_{2}} = \frac{2c_1}{K} - \frac{4c_{2}(c_1c_2 - c_3)}{K^{2}}$,
\item $\frac{\partial r_{12}}{\partial c_{3}} = \frac{-2}{K} - \frac{4c_{3}(c_1c_2 - c_3)}{K^{2}}$,
\item $\frac{\partial r_{22}}{\partial c_{1}} = \frac{-2c_1}{K} - \frac{2c_{1}(1 - c^{2}_1 + c^{2}_2 - c^{2}_3)}{K^{2}}$,
\item $\frac{\partial r_{22}}{\partial c_{2}} = \frac{2c_2}{K} - \frac{2c_{2}(1 - c^{2}_1 + c^{2}_2 - c^{2}_3)}{K^{2}}$,
\item $\frac{\partial r_{22}}{\partial c_{3}} = \frac{-2c_3}{K} - \frac{2c_{3}(1 - c^{2}_1 + c^{2}_2 - c^{2}_3)}{K^{2}}$,
\item $\frac{\partial r_{32}}{\partial c_{1}} = \frac{2}{K} - \frac{4c_{1}(c_1 + c_2c_3)}{K^{2}}$,
\item $\frac{\partial r_{32}}{\partial c_{2}} = \frac{2c_3}{K} - \frac{4c_{2}(c_1 + c_2c_3)}{K^{2}}$,
\item $\frac{\partial r_{32}}{\partial c_{3}} = \frac{2c_2}{K} - \frac{4c_{3}(c_1 + c_2c_3)}{K^{2}}$,
\end{itemize}

the entries of the matrices $\mathbf{A}_{3}$ are respectively,
\begin{itemize}
\item $\frac{\partial r_{13}}{\partial c_{1}} = \frac{2c_3}{K} - \frac{4c_{1}(c_2 + c_1c_3)}{K^{2}}$,
\item $\frac{\partial r_{13}}{\partial c_{2}} = \frac{2}{K} - \frac{4c_{2}(c_2 + c_1c_3)}{K^{2}}$,
\item $\frac{\partial r_{13}}{\partial c_{3}} = \frac{2c_1}{K} - \frac{4c_{3}(c_2 + c_1c_3)}{K^{2}}$,
\item $\frac{\partial r_{23}}{\partial c_{1}} = \frac{-2}{K} - \frac{4c_{1}(c_2c_3 - c_1)}{K^{2}}$,
\item $\frac{\partial r_{23}}{\partial c_{2}} = \frac{2c_3}{K} - \frac{4c_{2}(c_2c_3 - c_1)}{K^{2}}$,
\item $\frac{\partial r_{23}}{\partial c_{3}} = \frac{2c_2}{K} - \frac{4c_{3}(c_2c_3 - c_1)}{K^{2}}$,
\item $\frac{\partial r_{33}}{\partial c_{1}} = \frac{-2c_1}{K} - \frac{2c_{1}(1-c^{2}_1-c^{2}_2+c^{2}_3)}{K^{2}}$,
\item $\frac{\partial r_{33}}{\partial c_{2}} = \frac{-2c_2}{K} - \frac{2c_{2}(1-c^{2}_1-c^{2}_2+c^{2}_3)}{K^{2}}$,
\item $\frac{\partial r_{33}}{\partial c_{3}} = \frac{2c_3}{K} - \frac{2c_{3}(1-c^{2}_1-c^{2}_2+c^{2}_3)}{K^{2}}$,
\end{itemize}

\subsection{Derivation on Robust Weight Function Corresponding to the Tukey-Lambda Distribution}
\label{App: RWF of Tukey}

When the shape parameter $\lambda = 0$, the probability density function (pdf) of Tukey-Lamba distribution has the closed form as
\begin{equation}
P(x; \mu, k) = \frac{1}{k(e^{\frac{x - \mu}{2k}} + e^{-\frac{x - \mu}{2k}})^2},
\end{equation}
which is identical to the Logistic distribution. We assume $\mu = 0$ and thus the robust weight function is derived by
\begin{align}
\omega(x) &= -\frac{1}{2x}\frac{\partial \log{P(x;k)}}{\partial x} \\ \nonumber
		  &= \frac{1}{2kx}\frac{e^{\frac{x}{k}} - 1}{e^{\frac{x}{k}} + 1} \\ 
          &= \left\{ \begin{array}{ll} \frac{1}{2k^2(e^{\frac{x}{k}} + 1)}, \textup{if} \vert x\vert \leq \epsilon\\
                 \frac{e^{\frac{x}{k}} - 1}{2kx(e^{\frac{x}{k}} + 1)}, \textup{if} \vert x \vert > \epsilon
                 \end{array}, \right. \nonumber
\end{align}
where $\epsilon$ is a small positive number.

\section*{Acknowledgment}
The authors would like to thank Dr. Yi Yu for the careful proofreading, Dr. Guillermo Gallego, Dr. Yuchao Dai and Mr. Liu Liu for sharing their thoughts.

The research leading to these results is supported by the Australian Centre for Robotic Vision. The work is furthermore
supported by ARC grants DE150101365. Yi Zhou acknowledges the financial support from the China Scholarship Council for his PhD Scholarship No.201406020098.

\ifCLASSOPTIONcaptionsoff
  \newpage
\fi

\bibliographystyle{IEEEtran}
\bibliography{IEEEabrv,myBib}

\vspace{-1cm}
\vfill
\begin{IEEEbiography}
[{\includegraphics[width=1in,height=1.25in,clip,keepaspectratio]{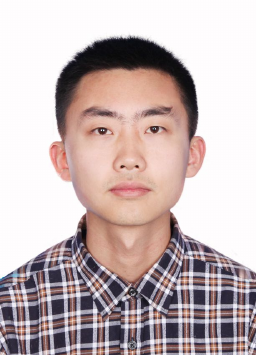}}]
{Yi Zhou} received the B. Sc. degree in Aircraft Manufacturing and Engineering from
Beijing University of Aeronautics and Astronautics (BUAA), China in 2012. He is currently a Ph. D. candidate in Research School of Engineering, the Australian National University (ANU). He was awarded the NCCR Fellowship Award for the research on event based vision in 2017 by the Swiss National Science Foundation through the National Center of Competence in Research (NCCR) Robotics. His research interests include Visual Odometry/SLAM (Simultaneous Localization and Mapping), geometry problems in computer vision and dynamic vision sensors.\\
E-mail: yi.zhou@anu.edu.au (Corresponding author)
\end{IEEEbiography}
\begin{IEEEbiography}
[{\includegraphics[width=1in,height=1.25in,clip,keepaspectratio]{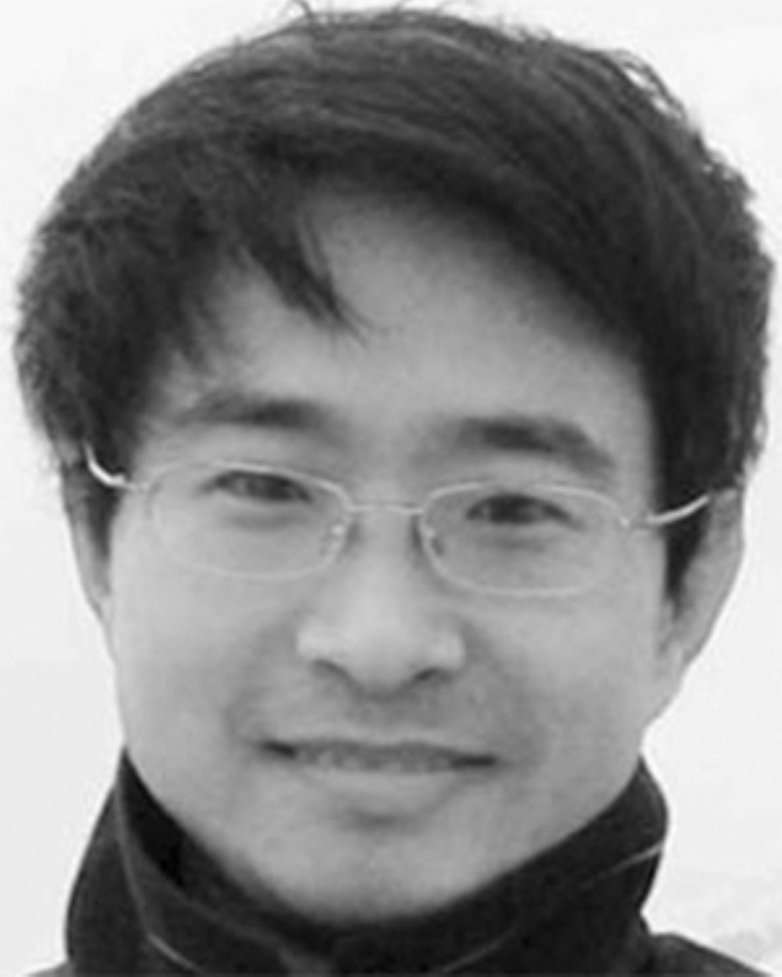}}]
{Hongdong Li} Hongdong Li is a Chief Investigator of the Australian ARC Centre of Excellence for Robotic Vision, Australian National University. His research interests include geometric computer vision, pattern recognition, computer graphics and combinatorial optimization. He is an Associate Editor for IEEE TPAMI and served as Area Chair for recent CVPR, ICCV and ECCV conferences. Jointly with coworkers, he has won a number of prestigious computer vision awards, including the CVPR Best Paper Award, the Marr Prize (Honorable Mention), the DSTO Best Fundamental Contribution to Image Processing Paper award and the best algorithm prize at the NRSFM Challenge at CVPR 2017. He is a program co-chair for ACCV 2018.\\
E-mail: hongdong.li@anu.edu.au
\end{IEEEbiography}

\begin{IEEEbiography}[{\includegraphics[width=1in,height=1.25in,clip,keepaspectratio]{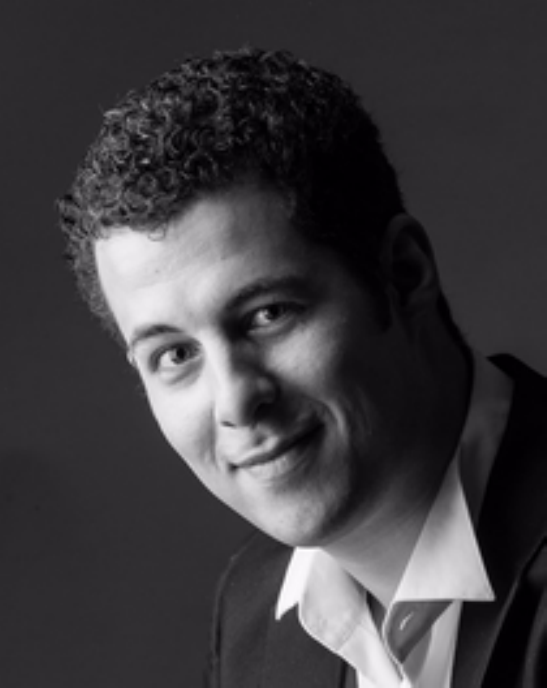}}]
{Laurent Kneip}
Laurent Kneip obtained a Dipl.-Ing. degree in mechatronics from Friedrich-Alexander University Erlangen-Nurnberg in 2009 and a PhD degree from ETH Zurich in 2013, in the fields of mobile robotics and computer vision. After his PhD, he was granted a Discovery Early-Career Researcher Award from the Australian Research Council and became a senior researcher at the Australian National University and a member of the ARC Center of Excellence for Robotic Vision. Since 2017, he has been an Assistant Professor at ShanghaiTech, where he founded and directs the Mobile Perception Lab. His most well-known research output, the open-source project OpenGV, summarises his contributions in geometric computer vision. In 2017, Laurent Kneip was awarded the Marr Prize best paper award (honourable mention).\\
E-mail: lkneip@shanghaitech.edu.cn (Corresponding author)
\end{IEEEbiography}

\end{document}